\documentclass[letterpaper]{article} 
\usepackage{aaai2026}  
\usepackage{times}  
\usepackage{helvet}  
\usepackage{courier}  
\usepackage[hyphens]{url}  
\usepackage{graphicx} 
\usepackage{natbib}  
\usepackage{caption} 
\pdfoutput=1
\usepackage{algorithm}
\usepackage{algorithmic}

\usepackage{subcaption}
\usepackage{amssymb}
\usepackage{amsmath}
\usepackage{multirow}
\usepackage{booktabs}
\usepackage{pifont}
\usepackage{textcomp}
\usepackage{xcolor}
\usepackage{makecell}
\usepackage{adjustbox}

\usepackage{newfloat}
\usepackage{listings}

\usepackage[labeled]{multibib}
\newcites{one}{Reference 1}
\newcites{two}{Reference 2}
\DeclareCaptionStyle{ruled}{labelfont=normalfont,labelsep=colon,strut=off} 
\floatstyle{ruled}
\newfloat{listing}{tb}{lst}{}
\floatname{listing}{Listing}
\title{VK-Det: Visual Knowledge Guided Prototype Learning for Open-Vocabulary Aerial Object Detection}
\author{
	Jianhang Yao\textsuperscript{\rm 1}, Yongbin Zheng\textsuperscript{\rm 1}\thanks{Corresponding author.}, Siqi Lu\textsuperscript{\rm 1}, Wanying Xu\textsuperscript{\rm 1}, Peng Sun\textsuperscript{\rm 1}
}
\affiliations{
    \textsuperscript{\rm 1}National University of Defense Technology,\\

    yaojianhang0422@163.com, yaojianhang23@nudt.edu.cn, zybnudt@nudt.edu.cn, \\
    lusiqi9803@163.com, wanying\textunderscore xu@nudt.edu.cn, sunpeng@nudt.edu.cn
}

\usepackage{bibentry}

\begin{document}
	
\maketitle
\begin{abstract}
	To identify objects beyond predefined categories, open-vocabulary aerial object detection (OVAD) leverages the zero-shot capabilities of visual-language models (VLMs) to generalize from base to novel categories. Existing approaches typically utilize self-learning mechanisms with weak text supervision to generate region-level pseudo-labels to align detectors with VLMs semantic spaces. However, text dependence induces semantic bias, restricting open-vocabulary expansion to text-specified concepts. We propose $\textbf{VK-Det}$, a $\textbf{V}$isual $\textbf{K}$nowledge-guided open-vocabulary object $\textbf{Det}$ection framework $\textit{without}$ extra supervision. First, we discover and leverage vision encoder's inherent informative region perception to attain fine-grained localization and adaptive distillation. Second, we introduce a novel prototype-aware pseudo-labeling strategy. It models inter-class decision boundaries through feature clustering and maps detection regions to latent categories via prototype matching. This enhances attention to novel objects while compensating for missing supervision. Extensive experiments show state-of-the-art performance, achieving 30.1 $\mathrm{mAP}^{N}$ on DIOR and 23.3 $\mathrm{mAP}^{N}$ on DOTA, outperforming even extra supervised methods.
\end{abstract}

\section{1 Introduction}
\begin{figure}[t]
	\centering
	\begin{subfigure}[a]{0.4\textwidth}
		\includegraphics[width=\linewidth]{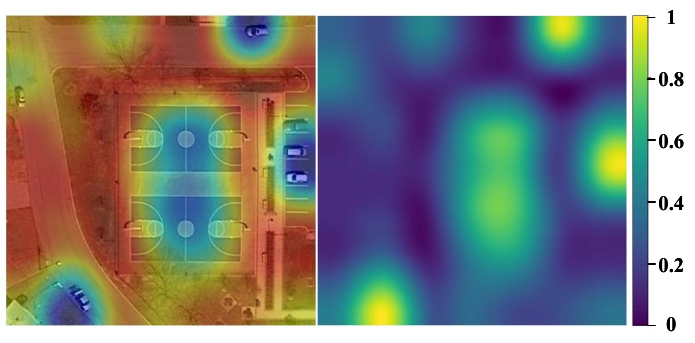}
		\caption{Informative region perception in VLMs.}
		\label{fig1a}
	\end{subfigure}
	\hfill 
	\begin{subfigure}[b]{0.4\textwidth}
		\includegraphics[width=\linewidth]{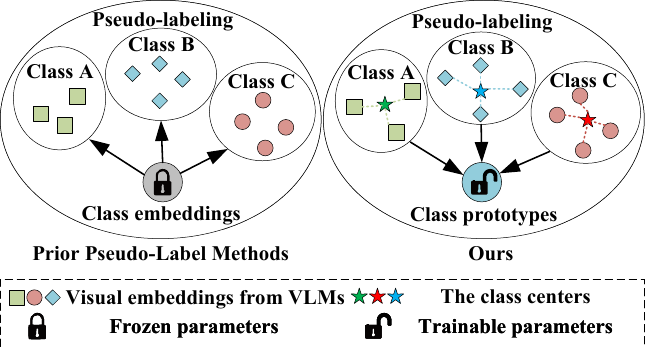}
		\caption{Comparison of our method with existing methods.}
		\label{fig1b}
	\end{subfigure}
	\caption{(a) visualizes the attention heatmap from the visual encoder of VLMs for an aerial image. The heatmap is derived by averaging multi-layer attentions. (b) compares our pseudo-labeling approach with conventional methods. }
	\label{fig1}
\end{figure}

Aerial object detection (AOD), which involves precisely localizing and classifying objects within aerial images, is essential for earth observation tasks including security monitoring, disaster response, and urban management \cite{1}. While deep learning has substantially improved AOD performance on closed-set benchmarks \cite{16}, existing methods remain limited to detecting only predefined object categories. To facilitate real-world deployment where countless unlabeled concepts exist, open-vocabulary aerial object detection (OVAD) is introduced to enable the recognition of novel objects \cite{5}.

Utilizing VLMs' zero-shot capabilities (e.g., RemoteCLIP \cite{7}), OVAD replaces trainable classifier weights with frozen semantic embeddings from the text encoder of VLMs. Current research on OVAD primarily focuses on knowledge distillation and pseudo-labeling:
\begin{itemize}
	\item Knowledge distillation transfers region-level semantic knowledge from VLMs to detectors, with ViLD \cite{8} pioneering the alignment between detector region features and cropped image embeddings from VLMs.
	\item Pseudo-labeling employs self-learning or external data to generate pseudo-labeled data, thereby expanding category coverage through extra supervision. For instance, CastDet \cite{5} employs a semi-supervised paradigm to produce high-quality pseudo-labels.
\end{itemize}

However, in AOD scenarios, knowledge distillation and pseudo-labeling underperform due to challenges in novel object localization, background interference, and textual noise, which ultimately lead to region-text misalignment. This compels us to confront an inherent issue: \textbf{whether we can automatically discover novel conceptual objects from visual knowledge while simultaneously maximizing their category boundaries for knowledge transfer}.

The visualization of attention heatmaps from VLMs' visual encoders reveals a notable phenomenon. This is evidenced by Fig. 1(\subref{fig1a}), where average attention maps across layers differentiate background from informative regions and assign higher weights to the latter without labels. Based on this observation, we design an adaptive selection mechanism and data augmentation method for small and elongated objects in aerial imagery. These components form the Adaptive Selection Knowledge Distillation (ASKD), enabling more efficient knowledge transfer. 
To better leverage informative regions, we propose a Prototype-Aware Pseudo-Label (PAPL) method. As shown in Fig. 1(\subref{fig1b}), unlike prior pseudo-label methods that classify class-agnostic image embeddings using frozen class embeddings \cite{13}, our method separates inter-class disparity via prototype learning and then generate trainable unknown class prototypes for classifier training. Finally, we propose the Synthetic Matching Inference (SMI) mechanism to evaluate the scores of novel classes through prototype matching and multi-level scoring. This approach integrates the distillation and prototype classifiers into a unified unknown category classifier, which works in conjunction with the localization network to assess the relevance of detected objects.

By combining the aforementioned modules, we introduce a novel framework termed \textbf{VK-Det}, a \textbf{Det}ector that relies only on \textbf{V}isual \textbf{K}nowledge for efficient knowledge distillation and pseudo-labeling optimization \textit{without additional supervised signals or data}. We argue that relying solely on visual knowledge from VLMs enables the detector to learn its semantic knowledge efficiently and achieve performance comparable to methods that utilize extra supervision. 

We evaluate VK-Det on two benchmark datasets, DIOR\cite{14} and DOTA\cite{1}. On DIOR, VK-Det achieves $30.1\%$ $\mathrm{mAP}$ on novel categories, outperforming state-of-the-art methods. Notably, previous methods rely on extra supervised signals, whereas VK-Det attains superior performance without extra supervision. On DOTA, VK-Det achieves a performance of $23.3\%$ $\mathrm{mAP}^{N}$ and $33.9\%$ $\mathrm{HM}$. Furthermore, our proposed PAPL method demonstrates superior performance and effectively mitigates the text illusion problem compared to approaches based on category-supervised pseudo-label generation. 

The \textbf{main contributions} of this paper are as follows:
\begin{itemize}
	\item  We identify the inherent capability of \textbf{informative regions perception} for potential objects in VLMs. Leveraging this insight, we design an ASKD framework to extract informative, region-level embeddings for more effective and adaptive distillation. 
	\item We introduce a PAPL approach that leverages prototype learning to generate high-quality pseudo-labels. Furthermore, we design prototype-based classifier and matching inference strategies to facilitate knowledge transfer.
	\item By integrating both components, we propose the VK-Det framework. Experiments on two standard benchmarks show that our method achieves state-of-the-art performance, surpassing even extra supervised methods.
\end{itemize}
\begin{figure*}[t]
	\centering
	\includegraphics[width=0.9\textwidth]{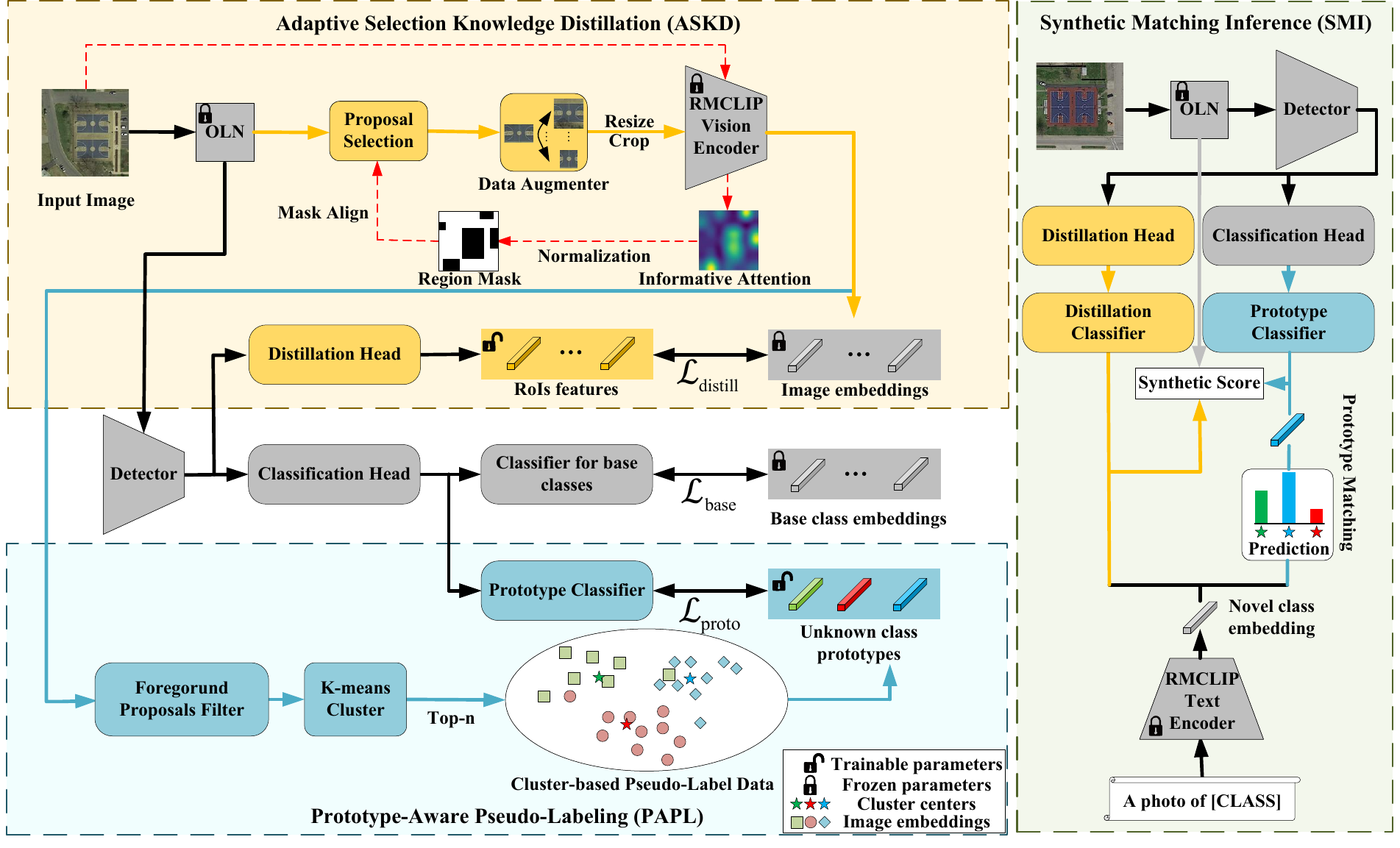} 
	\caption{The overall architecture of our VK-Det. By utilizing ASKD and PAPL in training, the detector enables comprehensive learning of unlabeled objects. During inference, SMI systematically evaluate novel category objects. In the figure, RMCLIP denotes RemoteCLIP\cite{7}.}
	\label{fig2}
\end{figure*}
\section{2 Related Work}
\textbf{Open-Vocabulary Object Detection.} 
The core idea of open-vocabulary object detection is to use joint visual-language modeling by leveraging pre-trained VLMs or image-text pairs as weakly supervised data to train a detector. Two dominant paradigms enable transferring image-level semantics from VLMs to region features: knowledge distillation and pseudo-labeling. Knowledge distillation extracts region-level semantic knowledge from pre-trained VLMs to empower detectors \cite{8,9,10,12}. Pseudo-labeling enhances the annotation quality of novel objects through additional supervision \cite{13,19}. Furthermore, current research on OVAD remains limited. DescReg \cite{21} leverages triplet loss to preserve visual similarity structures in the classification space and enhance knowledge transfer. CastDet \cite{5} employs a semi-supervised model with pseudo-labeled sequences to expand the class vocabulary. LAE-DINO \cite{22} uses additional data to train the detector by combining dynamic vocabulary construction with visually guided text prompts.

Unlike existing OVAD methods, our approach optimizes knowledge distillation for aerial imagery without extra data. To eliminate the "extra supervision" bias from pseudo labels, which are novel class labels or textual signals that appear during training, we propose a prototype-aware method. This method generates category prototypes and matches them dynamically, effectively suppressing noisy supervision and ensuring reliable semantic updates.

Further details on related work are in \textbf{Appendix A}.

\section{3 Methods}
\subsection{3.1 Preliminaries $\&$ Overview} 
\textbf{Preliminaries.} OVAD trains models on labeled data for base categories $C_B$, allowing them to localize and classify objects from novel categories $C_N$ during inference, where $C_B \cap C_N = \emptyset$. Its core principle involves implicitly learning semantic features of unlabeled objects in training images, enabling the alignment between visual and textual features in a unified embedding space during inference.

Currently, to address the open-world localization problem, pre-trained Region Proposal Network (RPN) \cite{15} or Object Localization Network (OLN) \cite{24} models typically serve as fundamental components for OVAD. For an input image, they generate class-agnostic object proposals $P$, which consist of three mutually exclusive subsets: foreground proposals $P_{fg}$ contain base-class $C_B$ objects; targeted proposals $P_{tg}$ contain unknown-class $C_U$ instances; and untargeted proposals $P_{ug}$ contain background-class $C^{BG}_{U}$ instances such as aerial imagery of woods or houses, with $P = P_{fg} \cup P_{tg} \cup P_{ug}$. Notably, $C_U \supset C_N$ and $C_B \cap C_U = \emptyset$. 

\textbf{Knowledge distillation} has emerged as an effective strategy for transferring semantic knowledge from VLMs to detectors. 
For each proposal ${p}\in{P}$, knowledge distillation crops the region, encodes it through the visual encoder of VLMs to generate image embeddings $v$, uses the detector simultaneously with a Region of Interest (RoI) extractor to extract $f_{\text{roi}}$, and aligns these representations by minimizing their orthogonal similarity. 
In inference, by generating text embeddings for novel categories $\{t^N_c \mid c \in \mathcal{C}_{N}\}$ via VLMs' text encoders, the category of detection boxes is inferred by computing the similarity probability between $f_{\text{roi}}$ of proposals and $t_{c}$, expressed as: 
\begin{small}
	\begin{equation}
	P(f_{roi}, t^N_c) = -\log{\frac{\exp(\langle f_{roi},t^N_{c}\rangle)}{\sum_{c\in\mathcal{C}_{N}}\exp(\langle f_{roi},t^N_{c}\rangle)}},
	\end{equation}
\end{small} where $\langle \cdot, \cdot \rangle $ presents their cosine similarity. 

\textbf{Pseudo-labeling} enhances novel category detection by assigning labels $C_N$ to proposals $P$ via similarity matching between proposals and novel class text embeddings $\{t^N_c\mid c\in\mathcal{C}_{N}\}$. This process builds high-confidence pseudo-annotated data. With this data, the detector learns robust semantic knowledge by jointly optimizing bounding box regression and classification for novel categories. 

\textbf{Overview.} In traditional distillation methods, distilling knowledge solely from $P$ introduces noisy background features and hampers the learning of category correlations due to the inaccurate localization of unknown-class proposals. This significantly undermines the efficiency of knowledge distillation. To obtain more informative image embeddings, we propose the ASKD module (Section 3.2). It addresses background interference and information destruction during knowledge extraction by exploiting VLMs’ ability to perceive informative regions in aerial images, enhancing feature alignment granularity. Moreover, current pseudo-labeling methods rely on additional supervised signals, inherently constraining open-vocabulary space expansion. Consequently, detection boundaries become limited by prior semantic knowledge, creating a category coverage bottleneck. To eliminate reliance on text embeddings, we propose an unsupervised pseudo-labeling method PAPL (Section 3.3) based on prototype learning. 
Finally, SMI (Section 3.4) is employed to integrate the outputs of the classifiers of ASKD and PAPL along with the localization network, thereby enabling a comprehensive determination of the existence probability of unknown category objects.

\subsection{3.2 Adaptive Selection Knowledge Distillation}
\textbf{Informative Region Perception.} VLMs exhibit strong zero-shot image-level classification capabilities, yet they struggle with fine-grained region localization. We observe that averaging attention maps across layers of the visual encoder assigns higher weights to informative regions. By leveraging spatial priors, our adaptive proposal selection module and data augmenter extract informative proposals $P_{inf} = P_{fg} \cup P_{tg}$ to facilitate efficient knowledge distillation. Unlike threshold-based and count-based filtering methods \cite{8}, our approach better preserves semantic correlations within VLMs' region embeddings. 

\textbf{Adaptive Proposal Selection.} To address the spatial mismatch between low-resolution attention maps and high-resolution images, we propose an attention normalization method inspired by \cite{35}. Specifically, we apply a scaling factor $\lambda$ and subsequently perform sigmoid activation to transform the original attention maps.
\begin{small}
	\begin{equation}
	\tilde{Attn}=\sigma(Attn\cdot\lambda)
	\label{eq:1} 
	\end{equation}
\end{small} where $\sigma(\cdot)$ denotes the sigmoid function. Subsequently, we integrate an adaptive shifting mechanism:
\begin{equation}
M = \tilde{Attn} + \max\left(1 - \mathbb{E}[\tilde{Attn}], 0\right)
\label{eq:attention_adjustment}
\end{equation}

This operation guarantees non-negative attention values while maintaining the distribution characteristics, thereby generating a normalized attention mask $M\in R^{H\times W}$. The corresponding region $R_i \subset M$ is extracted through bilinear interpolation for each proposal $p_i$. Subsequently, the regional average response is computed to select informative region proposals based on a predefined threshold criterion. Mathematically, this process can be formulated as:
\begin{small}
	\begin{equation}
	\tilde{w_i}=\mathbb{I}\left[\frac{1}{\mid R_i\mid}\sum_{(x,y)\in R_i}M(x,y)\geq1\right]
	\label{eq:2} 
	\end{equation}
\end{small}

This thresholding mechanism effectively distinguishes between informative regions ($\tilde{w}_i=1$) and non-informative regions ($\tilde{w}_i=0$), forming a subset of proposals $P_{inf}=\{p_i \mid \tilde{w}_i = 1\}$ that is optimized for the detection task.

\textbf{Data Augmenter Based on Max-Min Edge Jitter.} In VLMs, the non-adaptive cropping mechanisms of visual encoders significantly hinder the feature extraction process. Specifically, center-cropping objects with extreme aspect ratios results in the loss of crucial informative features, which compromises the alignment of semantic spaces during knowledge distillation. Moreover, given that aerial objects often exhibit local similarities across categories, the appropriate contextual receptive fields are essential for achieving high detection performance \cite{25}.

To address this, we propose an aspect ratio adaptive data augmenter to generate an enhanced proposal set $P_{aug}$. 
Given a proposal $\{p_i = (x_1, y_1, x_2, y_2)\mid{p_i} \in{P_{inf}}\}$ with width $w = x_2 - x_1$ and height $h = y_2 - y_1$, its aspect ratio is defined as $r = \max(w/h, h/w)$, reflecting its geometric characteristics. Proposals with $\log(r) > \alpha$ are classified as having extreme aspect ratios, where $\alpha$ is a pre-defined threshold. For such proposals, we define the dimensions as follows:
\begin{equation}
l=\max(w,h),\quad s=\min(w,h)
\end{equation}

We employ two distinct strategies to augment $P_{inf}$: Longer-side jittering perturbs $l$ within $\delta$ while fixing the maximum size. It can be expressed as:
\begin{small}
	\begin{equation}
	p_i^{\prime}=\left(c_x-\frac{l_\delta}{2},c_y-\frac{l_\delta}{2},c_x+\frac{l_\delta}{2},c_y+\frac{l_\delta}{2}\right),
	\end{equation}
	\begin{equation}
	l_\delta=l+\sigma \cdot s \cdot \epsilon,\quad\epsilon\sim\mathcal{N}(0, I),
	\end{equation} 
\end{small} where $c_x = \frac{x_1 + x_2}{2}, c_y = \frac{y_1 + y_2}{2}$ denote centroid coordinates. Set the variance jitter coefficient to a fixed value $\sigma$.

Shorter-side jittering proportionally scales $\delta$ while fixing the minimum size.
\begin{small}
	\begin{equation}
	p_i^{\prime}=\left(c_x-\frac{s_\delta}{2},c_y-\frac{s_\delta}{2},c_x+\frac{s_\delta}{2},c_y+\frac{s_\delta}{2}\right)
	\end{equation}
	\begin{equation}
	s_\delta=s+\sigma \cdot l \cdot \epsilon,\quad\epsilon\sim\mathcal{N}(0, I)
	\end{equation}
\end{small}

These augmented region proposals $P_{aug}$ enable the model to learn both local and global region views during training, thereby enhancing feature extraction for objects with extreme aspect ratios, such as aerial ships and bridges. 

\textbf{Loss Function.} Once $P_{aug}$ is obtained, informative region features can be extracted using the visual encoder of the VLMs. Distilling these features minimizes the distance of the feature space between the detector and the visual encoder of the VLMs.

Specifically, region features $f_{roi}(p_i^{\prime})$ are generated using the detector's RoI extractor. Meanwhile, the corresponding cropped image features $v(p_i^{\prime})$ are extracted using the visual encoder for the same proposal. The feature distillation is performed by minimizing the $L_1$ distance between two sets of features, thereby enforcing geometric consistency between region features and the VLMs' semantic space.
\begin{equation}
\mathcal{L}_{distill}=\frac{1}{| P_{aug} |}\sum_{i}\parallel f_{roi}(p_i^{\prime})-{v(p_i^{\prime})}\parallel_{1}
\label{eq:jitter_model}
\end{equation}

For training base classes, we follow \cite{8} by replacing the weights of the trainable classifiers with frozen text embeddings of base classes $\{t_c^B \mid c \in \mathcal{C}_{B}\}$, which are generated by the VLM’s text encoder. Additionally, we introduce a learnable background embedding $t_{bg}^B$, such that $t^B = \{t_c^B, t_{bg}^B\}$. For each class-agnostic proposal $p_i$ in the input image, the cross-entropy loss function is defined as: 
\begin{equation}
\mathcal{L}_{base}=\mathbb E _ {(f(p), t)} [P(f _ {roi}(p _ i), t ^ B)]
\label{eq:base_loss}
\end{equation}
\subsection{3.3 Prototype-Aware Pseudo-Labeling} 

\textbf{Unsupervised Pseudo-Labeled Data Generation.} Notably, $P_{aug}$ contains numerous unknown category objects $P_{tg}$. To generate high-quality pseudo-labeled data of unknown categories, we propose a PAPL method. It learns class decision boundaries from informative image embeddings using prototype learning. 

Specifically, $P_{aug}$ is processed by removing proposals that contain base categories $C_B$, based on RandBox's anchor position conditions \cite{26}, retaining only regions that potentially contain unknown categories. Their image embeddings undergo K-means clustering \cite{27} to capture inter-class differences among potential unknown categories, producing $k$ cluster centers $\{v_j\}_{j=1}^k$. To minimize intra-class noise, the top $n$ nearest neighbor embeddings to each center within the embedding space are selected. Their corresponding proposals form a clean pseudo-labeled dataset for unknown classes, with labels ranging from $\text{unknown-}1$ to $\text{unknown-}k$ (denoted as $\mathcal{C}_U = \{\text{unknown-}1, \dots, \text{unknown-}k\}$), corresponding to $k$ cluster centers $\{v_j\}_{j=1}^k$.

However, blindly increasing $k$ may scatter features from the same object category into different unknown class clusters due to feature variability introduced by local crop encoding. Training with only a subset of these datasets introduces bias in the detector's RoI feature representation and multi-scale feature selection for the affected categories. Therefore, the value of $k$ should be appropriately chosen. This ensures that the detector learns comprehensive features and achieves consistent bounding box localization for unknown categories, accurately capturing their open semantic knowledge \cite{29}.

\textbf{Trainable Class Prototype Setting.} Given the clean pseudo-labeled data, we propose learnable class prototypes to replace the frozen text embeddings of novel classes. The lack of extra priors in unsupervised pseudo-labeled data hinders precise category learning. To enable effective learning from the pseudo-labeled data, we introduce $k$ trainable class prototypes $\{u_c \mid c \in \mathcal{C}_{U}\}$, where each prototype corresponds to a specific unknown category ('$\text{unknown-}j$', $j = 1, 2, \dots, k$).

These class prototypes incorporate adaptive proposal embeddings for novel categories and include an additional learnable background prototype $u_{bg}$ to address the issue of background proposal misclassification within informative regions. Therefore, the complete set of class prototypes is defined as $u = \{u_c, u_{bg}\}$.

Although the detector does not have access to explicit semantic labels for the pseudo-labeled data, the training process encourages it to distinguish and utilize inter-class variations in visual features, which are encoded into the learnable class prototypes. For each proposal $p_i$ in the pseudo-labeled dataset, the cross-entropy loss used to optimize an extra prototype classifier is defined as:
\begin{equation}
\mathcal{L}_{proto}=\mathbb{E}_{(f(p),u)}{P(f_{roi}(p_i), u)}
\label{eq:proto_loss}
\end{equation}

This loss promotes alignment between VLMs' feature space and the detector's class prototypes. Selecting the top-$n$ proposal embeddings ensures tighter intra-class cohesion within the embedding space. As a result, the detector achieves improved discrimination of novel category features in informative regions. Further details on the prototype classifier are in \textbf{Appendix B}.\\

\begin{table*}[!t]
	\centering
	\renewcommand{\arraystretch}{0.8} 
	\small 
	\begin{tabular}{@{}lccrcccccccc@{}} 
		\toprule[0.8pt] 
		\multirow{2}{*}{\smash{\textbf{Method}}} & 
		\multirow{2}{*}{\smash{\textbf{Source}}} & 
		\multirow{2}{*}{\smash{\textbf{Backbone}}} & 
		\multirow{2}{*}{\smash{$\mathcal{S}_u$}} & 
		\multicolumn{4}{c}{\textbf{DIOR}} & 
		\multicolumn{4}{c}{\textbf{DOTA}} \\
		\cmidrule(r{2pt}){5-8} \cmidrule(l{2pt}){9-12}
		& & & & N & B & A & HM & N & B & A & HM \\ 
		\midrule[0.5pt] 
		RRFS$^{*}$ & CVPR22 & ResNet-101 & \ding{53} & 2.8 & \textcolor{gray}{41.9} & \textcolor{gray}{38.1} & 5.2 & 2.2 & \textcolor{gray}{47.1} & \textcolor{gray}{38.1} & 4.2 \\
		ContrastZSD$^{*}$ & TPAMI22 & ResNet-101 & \ding{53} & 3.9 & \textcolor{gray}{51.4} & \textcolor{gray}{41.9} & 7.2 & 2.8 & \textcolor{gray}{41.6} & \textcolor{gray}{33.8} & 5.2 \\
		ViLD$^{\text{\textdaggerdbl}}$ & ICLR22 & Resnet-50 & \ding{53} & 7.1 & \textcolor{gray}{63.5} & \textcolor{gray}{52.2} & 12.6 & 3.4 & \textcolor{gray}{63.8} & \textcolor{gray}{47.7} & 6.5 \\
		DescReg$^{*}$ & AAAI24 & ResNet-101 & \checkmark & 7.9 & \textcolor{gray}{68.7} & \textcolor{gray}{56.5} & 14.2 & 4.7 & \textcolor{gray}{68.7} & \textcolor{gray}{55.9} & 8.8 \\
		Castdet$^{\text{\textdaggerdbl}}$ & ECCV24 & ResNet-50 & \checkmark & \underline{29.8} & \textcolor{gray}{75.5} & \textcolor{gray}{66.5} & \textbf{42.7} & \underline{14.2} & \textcolor{gray}{64.3} & \textcolor{gray}{50.9} & \underline{23.3} \\
		Ours & - & ResNet-50 & \ding{53} & \textbf{30.1} & \textcolor{gray}{64.4} & \textcolor{gray}{57.5} & \underline{41.0} & \textbf{23.3} & \textcolor{gray}{62.0} & \textcolor{gray}{51.7} & \textbf{33.9}\\
		\bottomrule[0.8pt] 
	\end{tabular}
	\caption{Comparison with OVAD methods. where ${^\text{\textdaggerdbl}}$ represents that the results of our own implementation, under the same experimental setup as ours. $*$ represents results quoted from the original paper. $\mathcal{S}_u$ represents the incorporation of extra supervised signals from unknown categories during training. N, B, and A present $\mathrm{mAP}^{N}$, $\mathrm{mAP}^{B}$, and $\mathrm{mAP}^{A}$ respectively.}
	\label{table1}
\end{table*}

\begin{figure*}[t]
	\centering
	\begin{subfigure}[b]{0.33\textwidth}
		\centering
		\includegraphics[width=\linewidth]{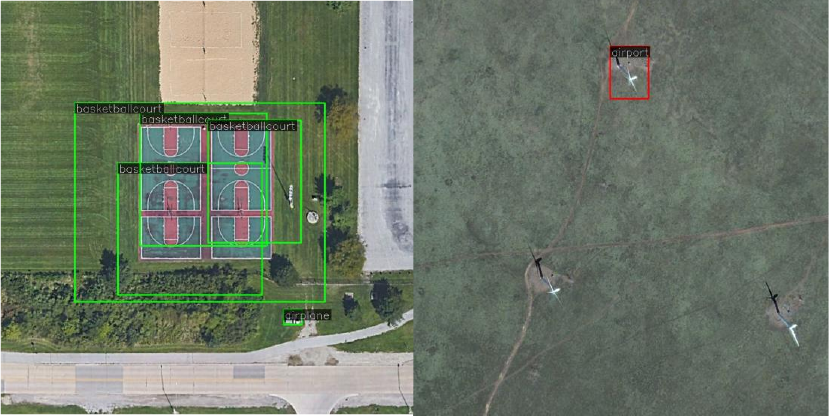}
		\caption{ViLD}
		\label{fig:sub1}
	\end{subfigure}
	\hfill
	\begin{subfigure}[b]{0.33\textwidth}
		\centering
		\includegraphics[width=\linewidth]{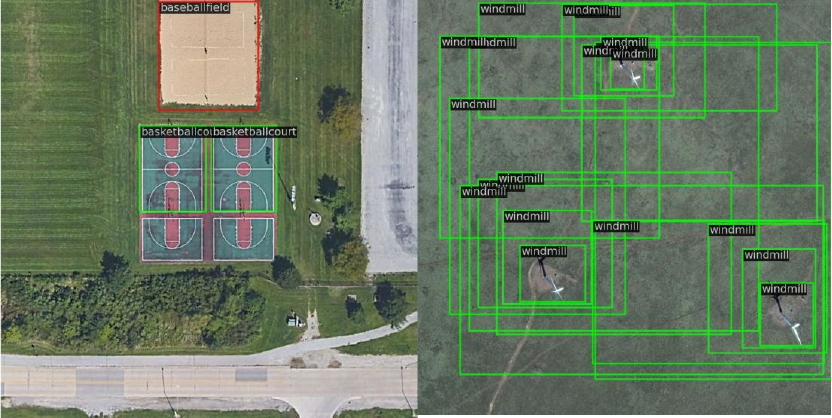}
		\caption{Castdet}
		\label{fig:sub2}
	\end{subfigure}
	\hfill
	\begin{subfigure}[b]{0.33\textwidth}
		\centering
		\includegraphics[width=\linewidth]{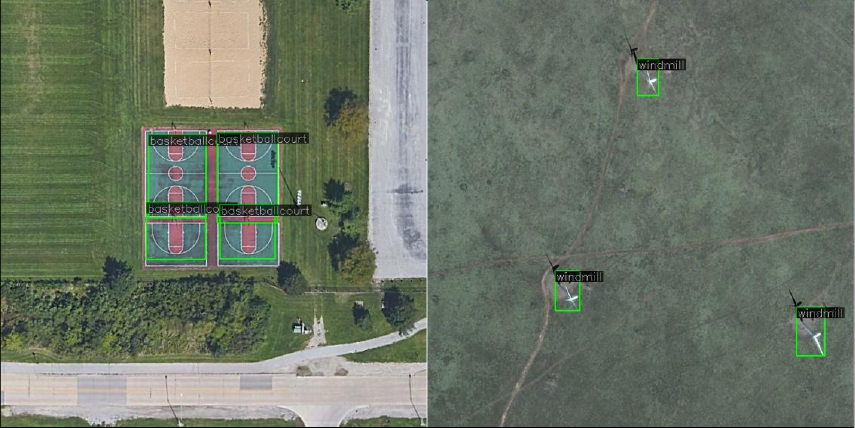}
		\caption{Ours}
		\label{fig:sub3}
	\end{subfigure}
	\caption{Visualization of open-vocabulary inference on DIOR dataset}
	\label{fig3}
\end{figure*}
\subsection{3.4 Synthetic Matching Inference}
Building upon traditional object detection frameworks such as Faster R-CNN \cite{15}, the two aforementioned methods enable the detector to learn generalized object representations beyond the distribution of the training data. 

To estimate confidence scores for novel objects, we introduce a SMI mechanism. Inspired by LP-OVOD \cite{19}, our framework aggregates distillation scores $Score_d$, prototype scores $Score_p$, and localization objectness $Score_l$ into a unified scoring scheme.

First, for each proposal $p$ generated by OLN, we utilize similarity scores modulated by a temperature parameter $\tau$ as one of the confidence metrics in the distillation head:
\begin{small}
	\begin{equation}
	Score_{d}=-\log{\frac{\exp(\langle f_{roi}(p),t^N_{c}\rangle /\tau)}{\sum_{c\in\mathcal{C}_{N}}\exp(\langle f_{roi}(p),t^N_{c}\rangle /\tau)}}
	\end{equation}
\end{small}

Additionally, regarding $Score_p$, matching text embeddings of novel categories to unknown class prototypes remains a challenging task.

To address it, our analysis indicates that a higher similarity between cluster centers and novel-category text embeddings corresponds to better alignment between object features and class prototypes. Therefore, for a given novel text embedding $t^N_c$, we identify its nearest neighbor cluster center $v_i$, referred to as 'unknown-i', among the cluster centers in orthogonal space, and directly select the corresponding prototype $\hat{u}_i$ for classification: 
\begin{small}
	\begin{equation}
	\hat{u}_i = \underset{j\in\{1,2,\dots,k\}}{\arg\max} \left(\langle t^N_c, v_j \rangle \right),
	\label{eq:cluster_assign}
	\end{equation} 
\end{small} where $\hat{u}_i$ denotes the $i$-th class prototype. 

Based on this selection, $Score_p$ is defined as:
\begin{small}
	\begin{equation}
	Score_p = -\log{\frac{\exp(\langle f_{roi}(p),\hat{u}_i\rangle /\tau)}{\sum_{j\in\{1,2,\dots,k\}}\exp(\langle f_{roi}(p),{u}_j\rangle /\tau)}}
	\label{eq:prototype_scoring}
	\end{equation}
\end{small}

The classification scores associated with these prototypes are aggregated using a dynamic weighting mechanism, ultimately forming the confidence estimate for unknown category objects within PAPL. The category score for a proposal in an image can then be expressed as:
\begin{equation}
Score_{cls}=\sqrt{Score_d \cdot Score_p}, 
\end{equation}

Furthermore, OLN, which is trained on base category data, also generates objectness scores ($Score_{l}$) based on the localization quality of object regions. These scores evaluate object confidence from a positional accuracy perspective. The synthetic score is formulated as:
\begin{equation}
Score_{s}=\sqrt{Score_{l} \cdot Score_{cls}}
\end{equation}

\section{4 Experiments}
\subsection{4.1 Experimental Setups}
\textbf{Datasets and Metrics.} 
To evaluate the effectiveness of VK-Det for OVAD, experiments are conducted on two established aerial benchmarks: DIOR and DOTA. Following established protocols \cite{21}, DIOR's categories are divided into 16 base categories and 4 novel categories, while DOTA's categories are split into 11 base categories and 4 novel categories. We conducted training on the training set and evaluated it on the validation set. The primary evaluation metric is mean Average Precision at an IoU threshold of 0.5. This includes base category performance ($\mathrm{mAP}^{B}$), novel category performance ($\mathrm{mAP}^{N}$), overall performance ($\mathrm{mAP}^{A}$), and harmonic mean ($\mathrm{HM}$), which balances detection capability between base and novel categories. Following Castdet \cite{5}, $\mathrm{HM}$ is calculated as:
\begin{small}
	\begin{equation}
	\mathrm{HM} = \frac{2 \cdot \mathrm{mAP}^{B} \cdot \mathrm{mAP}^{N}}{\mathrm{mAP}^{B} + \mathrm{mAP}^{N}}
	\end{equation}
\end{small}

Notably, $\mathrm{mAP}^{N}$ and $\mathrm{HM}$ are considered the primary evaluation metrics for OVAD on both datasets. Further details on dataset construction are in \textbf{Appendix D}.

\textbf{Implementation details.}
Our proposed method is implemented within the MMDetection toolbox \cite{30}. We use a Faster R-CNN \cite{15} with a ResNet-50 \cite{31} backbone as the detector. The pre-trained RemoteCLIP-ViT-B32 \cite{7} serves as the pre-trained VLM. The training process consists of two stages: In the first stage, the distillation head is trained via ASKD for 20 epochs with a batch size of 32 on a single A800 GPU, using SGD optimization (lr = 1e-3, weight decay = 1e-4). In the second stage, we select the top 500 proposals with the highest similarity to 20 cluster centers as pseudo-labels and establish 20 category prototypes for the detector. The detector is then fine-tuned using pseudo-labeled data for 12 epochs with a batch size of 64, during which the backbone and neck are frozen, while other settings from the first stage are retained.
\subsection{4.2 Comparison with State-of-the-Art Approaches}
We evaluate VK-Det against dedicated OVAD methods (CastDet \cite{5}, DescReg \cite{21}) and adapted general open-vocabulary detectors (RRFS \cite{32}, ContrastZSD \cite{33}, ViLD \cite{8}) for OVAD. 

To ensure a fair comparison, we preprocessed both datasets by filtering out training set images that contained annotations of novel categories, maintaining consistency with our methodology. We train ViLD through RemoteCLIP-based image-region feature distillation without textual supervision, while CastDet leverages extra supervision from novel classes to construct dynamic pseudo-labeled sequences that guide semantic learning. Further details on relevant model training are in \textbf{Appendix E}.

As summarized in Table {\ref{table1}}, our method demonstrates superior performance on both datasets compared to state-of-the-art open-vocabulary detectors. For instance, on DIOR, our method achieves $23.0\%$ higher $\mathrm{mAP}^{N}$ than ViLD (no extra supervision) and $0.3\%$ higher $\mathrm{mAP}^{N}$ than CastDet (extra supervision). This outperformance confirms our framework's effectiveness, establishing new SOTA. 

To intuitively evaluate the effectiveness of VK-Det, Fig.~$\ref{fig3}$ visualizes the detection results by comparing our detector with ViLD and CastDet methods. Correct novel class detections are shown in green, while erroneous detections appear in red. Our approach achieves precise novel class detection with minimal false positives. More results of the qualitative analysis are in \textbf{Appendix F}.

DOTA involves small objects and a large scale, which pose significant challenges. Our method achieves a $9.1\%$ improvement over the state-of-the-art, demonstrating strong generalization across diverse aerial scenarios.
\subsection{4.3 Ablation Study}
We conducted additional ablation studies on the DIOR dataset, including component-wise method analysis, ASKD ablation, PAPL ablation, and SMI ablation experiments.
\begin{table}[t]
	\renewcommand{\arraystretch}{0.8}
	\centering
	\small 
	\begin{tabular}{ccccccc}
		\toprule[0.8pt]
		ASKD & PAPL & SMI & N & B & A & HM \\
		\midrule[0.5pt] 
		\checkmark  & - & - & 7.8 & \textcolor{gray}{69.9} & \textcolor{gray}{57.5} & 14.0 \\
		\checkmark  & \checkmark & - & \underline{20.4} & \textcolor{gray}{68.0} & \textcolor{gray}{58.5} & \underline{31.4} \\
		\checkmark  & - & \checkmark & 20.1 & \textcolor{gray}{68.6} & \textcolor{gray}{58.9} & 31.1 \\
		\checkmark  & \checkmark & \checkmark & \textbf{30.1} & \textcolor{gray}{64.4} & \textcolor{gray}{57.5} & \textbf{41.0} \\
		\bottomrule[0.8pt]
	\end{tabular}
	\caption{Ablation study of the VK-Det framework.}
	\label{table2}
\end{table}

\textbf{Ablation study of the VK-Det framework.} Ablation studies evaluated core methods including ASKD, PAPL, and SMI in the VK-Det framework. Performance changes observed via stepwise integration in Table {\ref{table2}} demonstrate that ASKD boosted novel category detection, increasing $\mathrm{mAP}^{N}$ by $0.7\%$ compared to ViLD and confirming the critical role of information region perception in ASKD; subsequent optimization of the combining localization network score (excluding prototype classifier score) further improved novel category $\mathrm{mAP}^{N}$ to $20.1\%$, reflecting OLN's open-world localization characteristics; balancing ASKD and PAPL classifier weights then increased novel category $\mathrm{mAP}^{N}$ to $20.4\%$, indicating successful learning of category boundaries in PAPL; finally, full method collaboration achieved an optimal novel category $\mathrm{mAP}^{N}$ of $30.1\%$, demonstrating that these two methods acquire distinct and mutually complementary open semantic knowledge.

\textbf{Ablation study of the ASKD module.} To further validate ASKD's effectiveness, Table {\ref{table3}} explores refined modules in ASKD. "Mask" presents masked proposal selection, while "Enhancer" denotes our data augmenter. The results demonstrate the superiority of our proposed feature selection: Independently enhancing a subset of proposals yields a $3.2\%$ $\mathrm{mAP}^{N}$ gain over non-enhanced methods, while masked proposal selection improves performance by $4.5\%$ $\mathrm{mAP}^{N}$ compared to unmasked knowledge distillation. Combining both approaches further boosts detector performance.
\begin{table}[t]
	\renewcommand{\arraystretch}{0.8} 
	\centering
	\small 
	\begin{tabular}{ccccccc}
		\toprule[0.8pt]
		Mask & Enhancer & N & B & A & HM \\
		\midrule[0.5pt]
		- & - & 20.0 & \textcolor{gray}{64.5} & \textcolor{gray}{55.6} & 30.5 \\
		- & \checkmark & 23.2 & \textcolor{gray}{64.0} & \textcolor{gray}{55.8} & 34.1 \\
		\checkmark & - & \underline{24.5} & \textcolor{gray}{64.1} & \textcolor{gray}{56.2} & \underline{35.5} \\
		\checkmark & \checkmark & \textbf{30.1} & \textcolor{gray}{64.4} & \textcolor{gray}{57.5} & \textbf{41.0} \\
		\bottomrule[0.8pt]
	\end{tabular}
	\caption{Ablation study of the ASKD module.}
	\label{table3}
\end{table}

\begin{table}[t]
	\renewcommand{\arraystretch}{1.5}
	\centering
	\renewcommand{\arraystretch}{0.8} 
	\small 
	\begin{tabular}{ccccc}
		\toprule[0.8pt]
		- & N & B & A & HM \\
		\midrule[0.5pt]
		Extra supervision & \underline{28.1} & \textcolor{gray}{64.0} & \textcolor{gray}{56.8} & \underline{39.0} \\
		Ours & \textbf{30.1} & \textcolor{gray}{64.4} & \textcolor{gray}{57.5} & \textbf{41.0} \\
		\bottomrule[0.8pt]
	\end{tabular}
	\caption{Ablation study of the PAPL module.}
	\label{table4}
\end{table}
\textbf{Ablation study of the PAPL module.} Table {\ref{table4}} analyzes special pseudo-labeling cases by comparing our method with an extra supervised pseudo-labeling method for novel categories, where frozen text embeddings from fixed prompts generate quantitative pseudo-labels. Our approach outperforms the extra supervised pseudo-labeling method by $2.0\% \ \mathrm{mAP}^{N}$, as textual supervision induces hallucinations and noise in image regions, causing category bounding box shifts. Conversely, PAPL maps feature embeddings into a latent category space, enabling unknown category matching and significantly enhancing pseudo-label quality.

\textbf{Ablation study of the SMI module.} In the final inference stage, we conduct ablation studies to assess the impact of three scoring components: the distillation head score ($Score{_d}$), the classification head score ($Score{_p}$), and the localization network score ($Score{_l}$).

Table~{\ref{table5}} demonstrates that relying exclusively on distillation or classification scores traps the model in local optima at $7.8\%$ and $9.3\%$ $\mathrm{mAP}^{N}$, respectively. Merging two heads elevates $\mathrm{mAP}^{N}$ to $20.4\%$ through coordinated category discrimination, while integrating all three scores attains peak performance of $30.1\%$ $\mathrm{mAP}^{N}$.

\begin{figure}[t]
	\centering
	\begin{subfigure}[b]{0.45\textwidth}
		\centering
		\includegraphics[width=\linewidth]{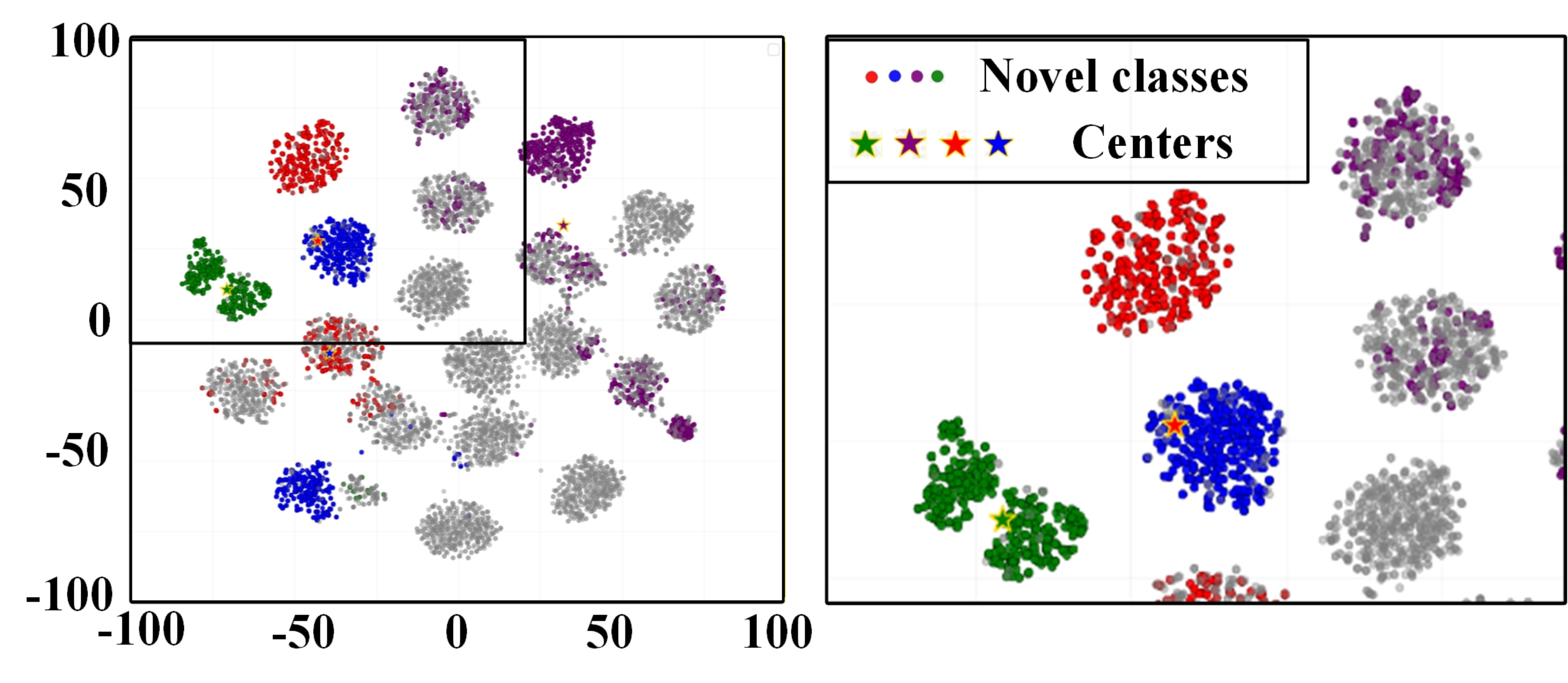}
		\caption{Pseudo-labels for overlapping with ground truth labels.}
		\label{fig1a}
	\end{subfigure}
	\hfill
	\begin{subfigure}[b]{0.45\textwidth}
		\centering
		\includegraphics[width=\linewidth]{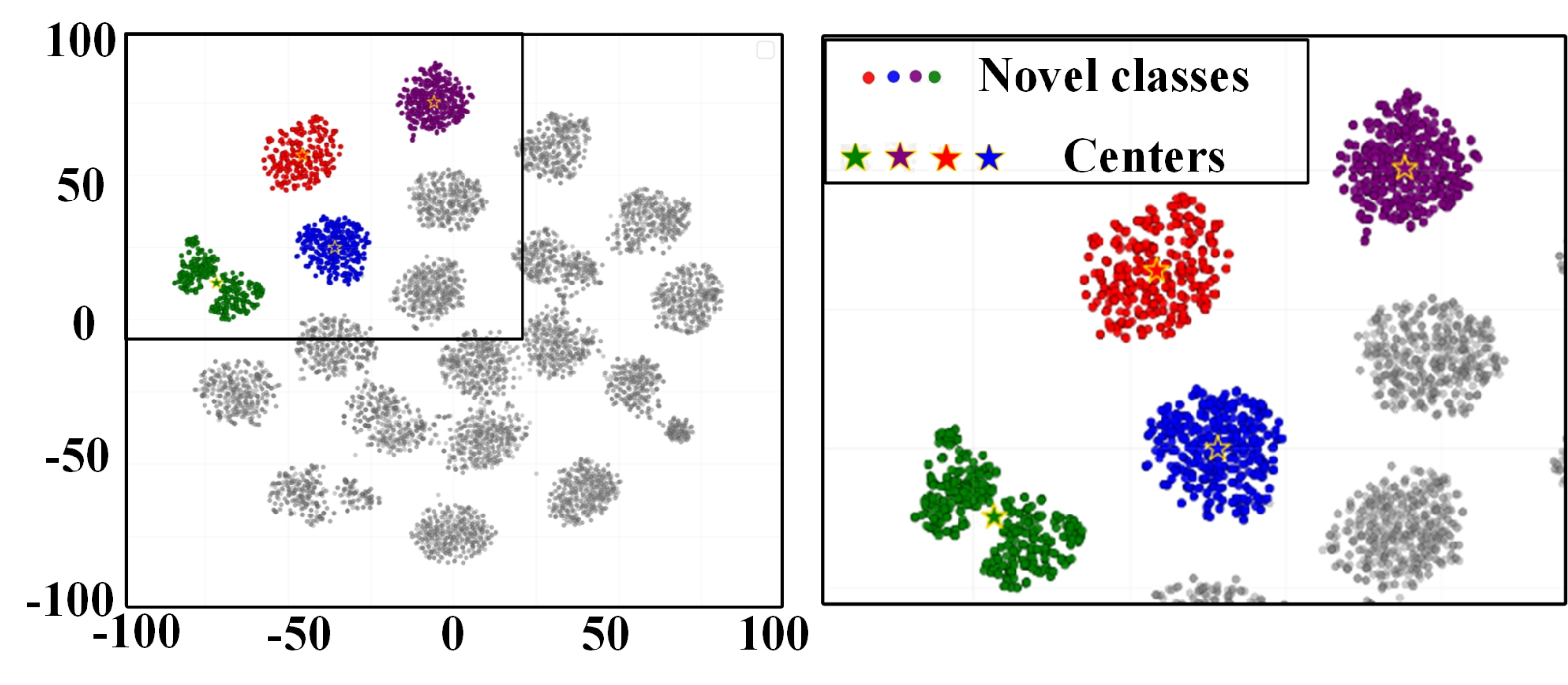}
		\caption{Pseudo-labels for text embedding selection.}
		\label{fig1b}
	\end{subfigure}
	\caption{Comparison of feature distributions: (a) Feature distribution in pseudo-labeled data with high IoU to the ground truth labels of novel classes; (b) Feature distribution in pseudo-labeled data exhibiting high similarity to the text embeddings of novel classes.}
	\label{fig4}
\end{figure}

Furthermore, t-SNE visualizations \cite{34} of novel category features demonstrate the efficacy of PAPL in Fig.~$\ref{fig4}$. The pseudo-labeled data generated by PAPL contains abundant novel category annotations; thus enabling the detector to learn distinguishable novel category features with minimal noise. This allows VK-Det to efficiently transfer novel category semantic knowledge from VLMs, significantly enhancing detection performance. More results of the ablation study can be found in \textbf{Appendix C}.

\begin{table}[t]
	\renewcommand{\arraystretch}{1.5}
	\centering
	\renewcommand{\arraystretch}{0.8} 
	\small 
	\begin{tabular}{cccccccc}
		\toprule[0.8pt]
		Score${_d}$ & Score${_p}$ & Score${_l}$ & N & B & A & HM \\
		\midrule[0.5pt]
		\checkmark  & - & - & 7.8 & \textcolor{gray}{69.9} & \textcolor{gray}{57.5} & 14.0 \\
		-  & \checkmark & - & 9.3 & \textcolor{gray}{69.6} & \textcolor{gray}{57.5} & 16.4 \\
		\checkmark  & \checkmark & - & 20.4 & \textcolor{gray}{68.0} & \textcolor{gray}{58.5} & 31.4 \\
		-  & \checkmark & \checkmark & 12.6 & \textcolor{gray}{64.4} & \textcolor{gray}{54.0} & 21.1 \\
		\checkmark  & - & \checkmark & \underline{24.8} & \textcolor{gray}{64.1} & \textcolor{gray}{56.2} & \underline{35.7} \\
		\checkmark  & \checkmark & \checkmark & \textbf{30.1} & \textcolor{gray}{64.4} & \textcolor{gray}{57.5} & \textbf{41.0} \\
		\bottomrule[0.8pt]
	\end{tabular}
	\caption{Ablation study of the SMI module.}
	\label{table5}
\end{table}
\section{5 Conclusion}
We propose to utilize informative region perception as guidance and construct a prototype learning-based classifier that efficiently and dynamically transfer knowledge from the visual encoder of VLMs. Unlike existing state-of-the-art methods that rely on extra supervision to generate pseudo-labels, we achieve efficient alignment of the detector feature space with the semantic space of VLMs through ASKD and PAPL, without additional data and supervision. Experiments show that when extra supervision is used to generate pseudo-labels, its robustness is lower than ours. This work could inspire further exploration of VLM visual knowledge spaces for dense prediction tasks. Future work will be devoted to developing efficient and lightweight methods for OVAD.

\bibliography{Formatting-Instructions-LaTeX-2026.bib}

\begin{thebibliography}{38}
\providecommand{\natexlab}[1]{#1}

\bibitem[{Cao et~al.(2019)Cao, Wei, Gaidon, Ar{\'e}chiga, and Ma}]{40}
Cao, K.; Wei, C.; Gaidon, A.; Ar{\'e}chiga, N.; and Ma, T. 2019.
\newblock Learning Imbalanced Datasets with Label-Distribution-Aware Margin
  Loss.
\newblock In \emph{Neural Information Processing Systems}.

\bibitem[{Chen et~al.(2019)Chen, Wang, Pang, Cao, Xiong, Li, Sun, Feng, Liu,
  Xu, Zhang, Cheng, Zhu, Cheng, Zhao, Li, Lu, Zhu, Wu, Liu, Dai, Wang, Shi,
  Ouyang, Loy, and Lin}]{30}
Chen, K.; Wang, J.; Pang, J.; Cao, Y.; Xiong, Y.; Li, X.; Sun, S.; Feng, W.;
  Liu, Z.; Xu, J.; Zhang, Z.; Cheng, D.; Zhu, C.; Cheng, T.; Zhao, Q.; Li, B.;
  Lu, X.; Zhu, R.; Wu, Y.; Liu, K.; Dai, J.; Wang, J.; Shi, J.; Ouyang, W.;
  Loy, C.~C.; and Lin, D. 2019.
\newblock MMDetection: Open MMLab Detection Toolbox and Benchmark.
\newblock \emph{ArXiv}, abs/1906.07155.

\bibitem[{Ding et~al.(2022)Ding, Xue, Xia, Bai, Yang, Yang, Belongie, Luo,
  Datcu, Pelillo, and Zhang}]{1}
Ding, J.; Xue, N.; Xia, G.-S.; Bai, X.; Yang, W.; Yang, M.~Y.; Belongie, S.;
  Luo, J.; Datcu, M.; Pelillo, M.; and Zhang, L. 2022.
\newblock Object Detection in Aerial Images: A Large-Scale Benchmark and
  Challenges.
\newblock \emph{IEEE Transactions on Pattern Analysis and Machine
  Intelligence}, 44(11): 7778–7796.

\bibitem[{Gu et~al.(2021)Gu, Lin, Kuo, and Cui}]{8}
Gu, X.; Lin, T.-Y.; Kuo, W.; and Cui, Y. 2021.
\newblock Open-vocabulary Object Detection via Vision and Language Knowledge
  Distillation.
\newblock In \emph{International Conference on Learning Representations}.

\bibitem[{He et~al.(2017)He, Gkioxari, Doll{\'a}r, and Girshick}]{35}
He, K.; Gkioxari, G.; Doll{\'a}r, P.; and Girshick, R.~B. 2017.
\newblock Mask R-CNN.

\bibitem[{He et~al.(2016)He, Zhang, Ren, and Sun}]{31}
He, K.; Zhang, X.; Ren, S.; and Sun, J. 2016.
\newblock Deep Residual Learning for Image Recognition.
\newblock In \emph{2016 IEEE Conference on Computer Vision and Pattern
  Recognition (CVPR)}, 770--778.

\bibitem[{Hu et~al.(2020)Hu, Jiang, Tang, Chen, Miao, and Zhang}]{39}
Hu, X.; Jiang, Y.; Tang, K.; Chen, J.; Miao, C.; and Zhang, H. 2020.
\newblock Learning to Segment the Tail.
\newblock \emph{2020 IEEE/CVF Conference on Computer Vision and Pattern
  Recognition (CVPR)}, 14042--14051.

\bibitem[{Huang et~al.(2022)Huang, Han, Cheng, and Zhang}]{32}
Huang, P.; Han, J.; Cheng, D.; and Zhang, D. 2022.
\newblock Robust Region Feature Synthesizer for Zero-Shot Object Detection.
\newblock In \emph{2022 IEEE/CVF Conference on Computer Vision and Pattern
  Recognition (CVPR)}, 7612--7621.

\bibitem[{Jia et~al.(2021)Jia, Yang, Xia, Chen, Parekh, Pham, Le, Sung, Li, and
  Duerig}]{41}
Jia, C.; Yang, Y.; Xia, Y.; Chen, Y.-T.; Parekh, Z.; Pham, H.; Le, Q.~V.; Sung,
  Y.-H.; Li, Z.; and Duerig, T. 2021.
\newblock Scaling Up Visual and Vision-Language Representation Learning With
  Noisy Text Supervision.
\newblock In \emph{International Conference on Machine Learning}.

\bibitem[{Kim et~al.(2021)Kim, Lin, Angelova, Kweon, and Kuo}]{24}
Kim, D.; Lin, T.-Y.; Angelova, A.; Kweon, I.-S.; and Kuo, W. 2021.
\newblock Learning Open-World Object Proposals without Learning to Classify.
\newblock \emph{IEEE Robotics and Automation Letters}, PP: 1--1.

\bibitem[{Li et~al.(2024{\natexlab{a}})Li, Zhang, Li, Li, Liu, Lin, and
  Li}]{29}
Li, J.; Zhang, J.; Li, J.; Li, G.; Liu, S.; Lin, L.; and Li, G.
  2024{\natexlab{a}}.
\newblock Learning Background Prompts to Discover Implicit Knowledge for Open
  Vocabulary Object Detection.
\newblock In \emph{2024 IEEE/CVF Conference on Computer Vision and Pattern
  Recognition (CVPR)}, 16678--16687.

\bibitem[{Li et~al.(2019)Li, Wan, Cheng, Meng, and Han}]{14}
Li, K.; Wan, G.; Cheng, G.; Meng, L.; and Han, J. 2019.
\newblock Object Detection in Optical Remote Sensing Images: A Survey and A New
  Benchmark.
\newblock \emph{ArXiv}, abs/1909.00133.

\bibitem[{Li et~al.(2023{\natexlab{a}})Li, Miao, Shi, Tan, Ren, Yang, and
  Pu}]{12}
Li, L.; Miao, J.; Shi, D.; Tan, W.; Ren, Y.; Yang, Y.; and Pu, S.
  2023{\natexlab{a}}.
\newblock Distilling DETR with Visual-Linguistic Knowledge for Open-Vocabulary
  Object Detection.
\newblock In \emph{2023 IEEE/CVF International Conference on Computer Vision
  (ICCV)}, 6478--6487.

\bibitem[{Li et~al.(2023{\natexlab{b}})Li, Guo, Yang, Liao, He, Zhou, and
  Yu}]{5}
Li, Y.; Guo, W.; Yang, X.; Liao, N.; He, D.; Zhou, J.; and Yu, W.
  2023{\natexlab{b}}.
\newblock CastDet: Toward Open Vocabulary Aerial Object Detection with
  CLIP-Activated Student-Teacher Learning.
\newblock In \emph{European Conference on Computer Vision}.

\bibitem[{Li et~al.(2024{\natexlab{b}})Li, Li, Dai, Hou, Liu, Liu, Cheng, and
  Yang}]{25}
Li, Y.; Li, X.; Dai, Y.; Hou, Q.; Liu, L.; Liu, Y.; Cheng, M.-M.; and Yang, J.
  2024{\natexlab{b}}.
\newblock LSKNet: A Foundation Lightweight Backbone for Remote Sensing.
\newblock \emph{ArXiv}, abs/2403.11735.

\bibitem[{Linderman et~al.(2017)Linderman, Rachh, Hoskins, Steinerberger, and
  Kluger}]{34}
Linderman, G.~C.; Rachh, M.; Hoskins, J.~G.; Steinerberger, S.; and Kluger, Y.
  2017.
\newblock Fast Interpolation-based t-SNE for Improved Visualization of
  Single-Cell RNA-Seq Data.
\newblock \emph{Nature methods}, 16: 243 -- 245.

\bibitem[{Liu et~al.(2023)Liu, Chen, Guan, Zhou, Zhu, and Zhou}]{7}
Liu, F.; Chen, D.; Guan, Z.-R.; Zhou, X.; Zhu, J.; and Zhou, J. 2023.
\newblock RemoteCLIP: A Vision Language Foundation Model for Remote Sensing.
\newblock \emph{IEEE Transactions on Geoscience and Remote Sensing}, 62: 1--16.

\bibitem[{Liu et~al.(2020)Liu, Gao, Sun, and Fang}]{3}
Liu, Z.; Gao, G.; Sun, L.; and Fang, Z. 2020.
\newblock HRDNet: High-Resolution Detection Network for Small Objects.
\newblock \emph{2021 IEEE International Conference on Multimedia and Expo
  (ICME)}, 1--6.

\bibitem[{Ma et~al.(2022)Ma, Luo, Gao, Li, Chen, Wang, Zhang, and Hu}]{10}
Ma, Z.; Luo, G.; Gao, J.; Li, L.; Chen, Y.; Wang, S.; Zhang, C.; and Hu, W.
  2022.
\newblock Open-Vocabulary One-Stage Detection with Hierarchical Visual-Language
  Knowledge Distillation.
\newblock \emph{2022 IEEE/CVF Conference on Computer Vision and Pattern
  Recognition (CVPR)}, 14054--14063.

\bibitem[{Pan et~al.(2024)Pan, Liu, Fu, Ma, Li, Paudel, van Gool, and
  Huang}]{22}
Pan, J.; Liu, Y.; Fu, Y.; Ma, M.; Li, J.; Paudel, D.~P.; van Gool, L.; and
  Huang, X. 2024.
\newblock Locate Anything on Earth: Advancing Open-Vocabulary Object Detection
  for Remote Sensing Community.
\newblock In \emph{AAAI Conference on Artificial Intelligence}.

\bibitem[{Pham, Vu, and Nguyen(2023)}]{19}
Pham, C.; Vu, T.; and Nguyen, K. 2023.
\newblock LP-OVOD: Open-Vocabulary Object Detection by Linear Probing.
\newblock \emph{2024 IEEE/CVF Winter Conference on Applications of Computer
  Vision (WACV)}, 768--777.

\bibitem[{Radford et~al.(2021)Radford, Kim, Hallacy, Ramesh, Goh, Agarwal,
  Sastry, Askell, Mishkin, Clark, Krueger, and Sutskever}]{6}
Radford, A.; Kim, J.~W.; Hallacy, C.; Ramesh, A.; Goh, G.; Agarwal, S.; Sastry,
  G.; Askell, A.; Mishkin, P.; Clark, J.; Krueger, G.; and Sutskever, I. 2021.
\newblock Learning Transferable Visual Models From Natural Language
  Supervision.
\newblock In \emph{International Conference on Machine Learning}.

\bibitem[{Redmon et~al.(2015)Redmon, Divvala, Girshick, and Farhadi}]{16}
Redmon, J.; Divvala, S.~K.; Girshick, R.~B.; and Farhadi, A. 2015.
\newblock You Only Look Once: Unified, Real-Time Object Detection.
\newblock \emph{2016 IEEE Conference on Computer Vision and Pattern Recognition
  (CVPR)}, 779--788.

\bibitem[{Redmon and Farhadi(2016)}]{36}
Redmon, J.; and Farhadi, A. 2016.
\newblock YOLO9000: Better, Faster, Stronger.
\newblock \emph{2017 IEEE Conference on Computer Vision and Pattern Recognition
  (CVPR)}, 6517--6525.

\bibitem[{Ren et~al.(2015)Ren, He, Girshick, and Sun}]{15}
Ren, S.; He, K.; Girshick, R.~B.; and Sun, J. 2015.
\newblock Faster R-CNN: Towards Real-Time Object Detection with Region Proposal
  Networks.
\newblock \emph{IEEE Transactions on Pattern Analysis and Machine
  Intelligence}, 39: 1137--1149.

\bibitem[{Singh et~al.(2017)Singh, Li, Sharma, and Davis}]{37}
Singh, B.; Li, H.; Sharma, A.; and Davis, L.~S. 2017.
\newblock R-FCN-3000 at 30fps: Decoupling Detection and Classification.
\newblock \emph{2018 IEEE/CVF Conference on Computer Vision and Pattern
  Recognition}, 1081--1090.

\bibitem[{Sun et~al.(2023)Sun, Fang, Wu, Wang, and Cao}]{42}
Sun, Q.; Fang, Y.; Wu, L.~Y.; Wang, X.; and Cao, Y. 2023.
\newblock EVA-CLIP: Improved Training Techniques for CLIP at Scale.
\newblock \emph{ArXiv}, abs/2303.15389.

\bibitem[{Vaze et~al.(2022)Vaze, Han, Vedaldi, and Zisserman}]{27}
Vaze, S.; Han, K.; Vedaldi, A.; and Zisserman, A. 2022.
\newblock Generalized Category Discovery.
\newblock \emph{2022 IEEE/CVF Conference on Computer Vision and Pattern
  Recognition (CVPR)}, 7482--7491.

\bibitem[{Wang et~al.(2023{\natexlab{a}})Wang, Liu, Du, Ding, Liao, Qi, Chen,
  and Liu}]{9}
Wang, L.; Liu, Y.; Du, P.; Ding, Z.; Liao, Y.; Qi, Q.; Chen, B.; and Liu, S.
  2023{\natexlab{a}}.
\newblock Object-Aware Distillation Pyramid for Open-Vocabulary Object
  Detection.
\newblock \emph{2023 IEEE/CVF Conference on Computer Vision and Pattern
  Recognition (CVPR)}, 11186--11196.

\bibitem[{Wang et~al.(2023{\natexlab{b}})Wang, Yue, Hua, and Zhang}]{26}
Wang, Y.; Yue, Z.; Hua, X.; and Zhang, H. 2023{\natexlab{b}}.
\newblock Random Boxes Are Open-world Object Detectors.
\newblock \emph{2023 IEEE/CVF International Conference on Computer Vision
  (ICCV)}, 6210--6220.

\bibitem[{Wang et~al.(2023{\natexlab{c}})Wang, Prabha, Huang, Wu, and
  Rajagopal}]{44}
Wang, Z.; Prabha, R.; Huang, T.; Wu, J.; and Rajagopal, R. 2023{\natexlab{c}}.
\newblock SkyScript: A Large and Semantically Diverse Vision-Language Dataset
  for Remote Sensing.
\newblock \emph{ArXiv}, abs/2312.12856.

\bibitem[{Yan et~al.(2021)Yan, Chang, Luo, Liu, Zhang, and Zheng}]{33}
Yan, C.; Chang, X.; Luo, M.; Liu, H.; Zhang, X.; and Zheng, Q. 2021.
\newblock Semantics-Guided Contrastive Network for Zero-Shot Object Detection.
\newblock \emph{IEEE Transactions on Pattern Analysis and Machine
  Intelligence}, 46: 1530--1544.

\bibitem[{Yang, Wu, and Chen(2019)}]{38}
Yang, H.; Wu, H.-Y.; and Chen, H. 2019.
\newblock Detecting 11K Classes: Large Scale Object Detection Without
  Fine-Grained Bounding Boxes.
\newblock \emph{2019 IEEE/CVF International Conference on Computer Vision
  (ICCV)}, 9804--9812.

\bibitem[{Zang et~al.(2024)Zang, Lin, Tang, Wang, and Lv}]{21}
Zang, Z.; Lin, C.; Tang, C.; Wang, T.; and Lv, J. 2024.
\newblock Zero-Shot Aerial Object Detection with Visual Description
  Regularization.
\newblock \emph{ArXiv}, abs/2402.18233.

\bibitem[{Zhai et~al.(2023)Zhai, Mustafa, Kolesnikov, and Beyer}]{43}
Zhai, X.; Mustafa, B.; Kolesnikov, A.; and Beyer, L. 2023.
\newblock Sigmoid Loss for Language Image Pre-Training.
\newblock \emph{2023 IEEE/CVF International Conference on Computer Vision
  (ICCV)}, 11941--11952.

\bibitem[{Zhang et~al.(2023)Zhang, Lei, Xie, Fang, Li, and Du}]{2}
Zhang, J.; Lei, J.; Xie, W.; Fang, Z.; Li, Y.; and Du, Q. 2023.
\newblock SuperYOLO: Super Resolution Assisted Object Detection in Multimodal
  Remote Sensing Imagery.
\newblock \emph{IEEE Transactions on Geoscience and Remote Sensing}, 61:
  1–15.

\bibitem[{Zhao et~al.(2022)Zhao, Zhang, Schulter, Zhao, B.G, Stathopoulos,
  Chandraker, and Metaxas}]{13}
Zhao, S.; Zhang, Z.; Schulter, S.; Zhao, L.; B.G, V.~K.; Stathopoulos, A.;
  Chandraker, M.; and Metaxas, D.~N. 2022.
\newblock Exploiting Unlabeled Data with Vision and Language Models for Object
  Detection.
\newblock \emph{ArXiv}, abs/2207.08954.

\bibitem[{Zhu and Chen(2023)}]{4}
Zhu, C.; and Chen, L. 2023.
\newblock A Survey on Open-Vocabulary Detection and Segmentation: Past,
  Present, and Future.
\newblock \emph{IEEE Transactions on Pattern Analysis and Machine
  Intelligence}, 46: 8954--8975.

\end{thebibliography}

\appendix          
\newpage           
\begin{center}
	\LARGE\textbf{Appendix}
\end{center}

\section{Overview}
\begin{figure*}[t]
	\centering
	\includegraphics[width=1\textwidth]{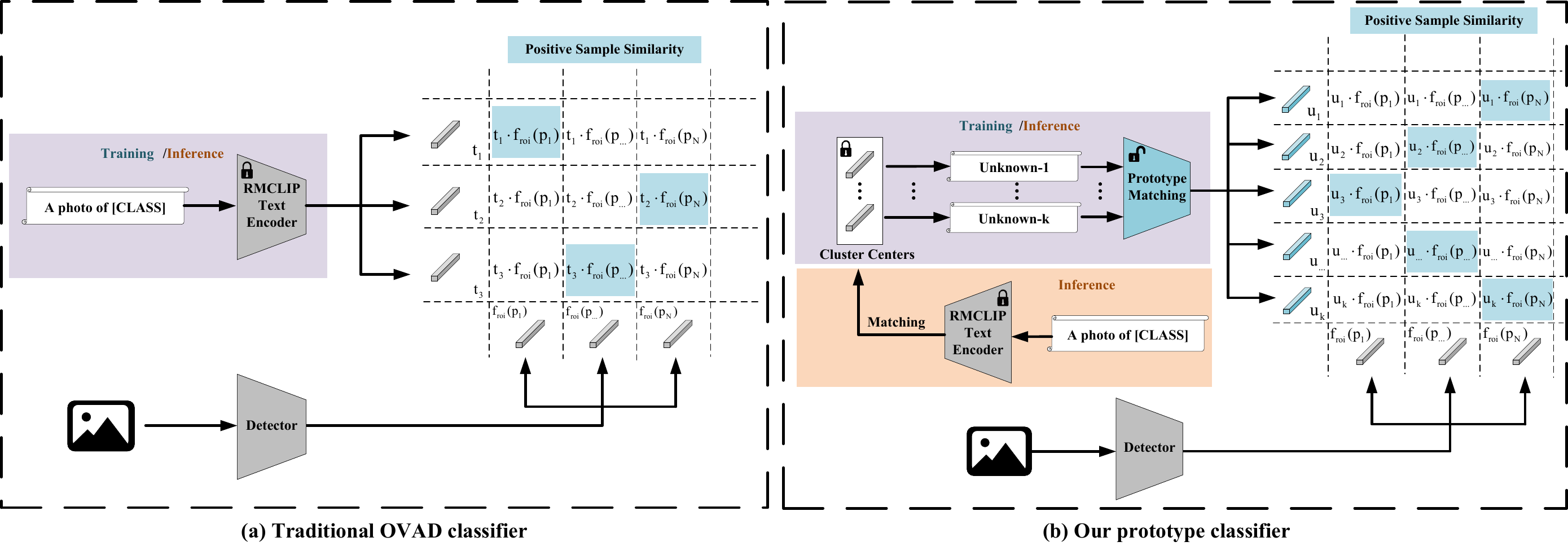} 
	\caption{Workflow of how to classify the pseudo-labeled data.}
	\label{fig5}
\end{figure*}

Due to space limitations, full details of the VK-Det study and additional experimental validation results cannot be included in the main text. Therefore, we provide comprehensive supplementary material for further discussion. These supplementary materials are organized into six sections. 
\textbf{Section A} provides a detailed discussion of supplementary related work not included in the main text, with a focus on closed-set object detection and pre-trained vision-language models. \textbf{Section B} elaborates on the workflow of the prototype classifier and its differentiation from traditional open-vocabulary object detection classifiers. \textbf{Section C} details ablation experiments that analyze parameter sensitivity in the VK-Det model. \textbf{Section D} details the construction methodology for the open-vocabulary aerial object detection dataset. \textbf{Section E} describes the training details of the relevant models. Finally, \textbf{Section F} presents the results of the qualitative analysis.

\section{A More Related Work}
\textbf{Closed-Set Object Detection.}
Closed-set object detection trains on a defined category-based dataset and evaluates on a validation set from the same distribution. Deep neural networks with top-down designs are commonly used to enhance accuracy.
To improve detection in open-vocabulary scenarios, researchers typically use large-vocabulary strategies like category-level semantic expansion \cite{36,37} and weakly supervised learning \cite{38} to increase recognizable categories.
Moreover, the long-tail distribution in open-vocabulary scenarios matches real-world category distributions. Consequently, long-tail object detection has been proposed, leveraging techniques such as data resampling \cite{39} and loss reweighting \cite{40} to mitigate the associated challenges.
However, such detection methods are often limited by small dataset sizes and imbalanced class distributions, which restrict category generalization and hinder their effectiveness in real-world applications.

In contrast, by utilizing the zero-shot recognition capabilities of VLMs, OVAD is better suited for AOD tasks demanding high open-vocabulary recognition performance.

\textbf{Pre-trained Vision-Language Models.}
VLMs pre-training is a learning paradigm that leverages large-scale image-text pairs to learn broad semantic associations between visual and textual modalities. This enables zero-shot prediction across diverse visual tasks. CLIP, a representative model, aligns matching image-text pairs in the embedding space while separating unrelated ones through contrastive learning. 
Inspired by CLIP, a series of subsequent studies have focused on optimizing the alignment quality between images and text within a unified embedding space, thereby further enhancing the model's zero-shot recognition capabilities \cite{41,42,43}.
To improve the adaptability of VLMs in aerial applications, researchers typically fine-tune multimodal encoders using domain-specific data, as exemplified by representative approaches such as RemoteCLIP \cite{7} and SkyCLIP \cite{44}. 

Building on RemoteCLIP's demonstrated zero-shot recognition capabilities in aerial imagery, we develop an open-vocabulary aerial object detector that effectively leverages VLMs' prior knowledge for transfer learning in downstream detection tasks.

\section{B Details about the Prototype Classifier Implementation}
As shown in Fig.~\ref{fig5}, the implementation of the prototype classifier classification loss begins by quantifying the similarity between unknown category labels and region features within the feature space. This is accomplished by constructing a score matrix which forms the basis for computing the cross-entropy loss. Specifically, the numerator of the loss function is composed of the similarity scores between the prototypes of unknown category labels and the features of positive samples in the pseudo-labeled data. The denominator is obtained by summing the exponentiated similarity scores across all sample pairs, including both positive and negative samples. This design enables the model to automatically learn discriminative features of different unknown categories from the pseudo-labeled data by maximizing the similarity of positive samples while minimizing the contribution of negative samples.

Notably, since our framework is capable of learning categories outside the closed-set of the training data through the prototype classifier, it involves a significantly larger number of unknown classes than the novel categories used in the final inference (e.g., urban road, housing estate). This is in contrast to traditional OVAD classifiers \cite{1}, which typically rely on frozen text embeddings as classifier weights. These text embeddings reflect real label features and are aligned with visual features in a shared embedding space. However, this paradigm fundamentally differs from our approach. In traditional methods, the region features extracted by the detector are often dissimilar to the frozen text embeddings, resulting in a score matrix with excessively low similarity. This severely hinders the optimization of the score matrix. Therefore, our work achieves a key breakthrough by designing a prototype-based classifier that enables the detector to effectively learn features of unknown classes beyond the distribution of the training labels.

\section{C Parameter Sensitivity Analysis}

\subsection{C.1 Selection of the Number of Cluster Centers}
\begin{table}[t]
	\renewcommand{\arraystretch}{1.5}
	\centering
	\renewcommand{\arraystretch}{0.8}
	\small
	\begin{tabular}{ccccc}
		\toprule[0.8pt]
		k & N & B & A & HM \\
		\midrule[0.5pt]
		10 & 28.7 & \textcolor{gray}{64.2} & \textcolor{gray}{57.1} & 39.7 \\
		20 & \textbf{30.1} & \textcolor{gray}{64.4} & \textcolor{gray}{57.5} & \textbf{41.0} \\
		30 & \underline{29.0} & \textcolor{gray}{64.4} & \textcolor{gray}{57.3} & \underline{40.0} \\
		\bottomrule[0.8pt]
	\end{tabular}
	\caption{Study on the number of cluster centers.}
	\label{table2}
\end{table}
As shown in Table 1, the comparison of model performance across different numbers of cluster centers indicates that performance plateaus when the number of cluster centers reaches $k=20$. It can be inferred that too few cluster centers ($k=10$) may result in inadequate feature aggregation for unknown categories, whereas too many cluster centers ($k=30$) may lead to over-dispersion of category features. This can cause instability in the prototype classifier and increased sensitivity to regional feature variations.

\subsection{C.2 Analysis of Foreground Proposal Filtering}
\begin{table}[t]
	\renewcommand{\arraystretch}{1.5}
	\centering
	\renewcommand{\arraystretch}{0.8} 
	\small 
	\begin{tabular}{ccccc}
		\toprule[0.8pt]
		- & N & B & A & HM \\
		\midrule[0.5pt]
		No filtering & \underline{21.4} & \textcolor{gray}{64.3} & \textcolor{gray}{55.7} & \underline{32.1} \\
		Ours & \textbf{30.1} & \textcolor{gray}{64.4} & \textcolor{gray}{57.5} & \textbf{41.0} \\
		\bottomrule[0.8pt]
	\end{tabular}
	\caption{The importance of foreground proposal filtering.}
	\label{table3}
\end{table}
Table 2 presents the effect of the foreground proposal filtering mechanism within the PAPL module. It can be observed that retaining foreground proposals results in a significant decrease in detection performance, with a reduction of $8.7\% \text{mAP}^N$. This performance drop is attributed to the interference caused by the clustering process of foreground proposal image embeddings with the feature distributions of unknown categories. As a result, the generated pseudo-labeled data contains considerable noise.
\subsection{C.3 Number of Class Prototypes Selected via Dynamic Integration during Inference}

\begin{table}[t]
	\renewcommand{\arraystretch}{1.5}
	\centering
	\renewcommand{\arraystretch}{0.8} 
	\small 
	\begin{tabular}{ccccc}
		\toprule[0.8pt]
		\# prototypes & N & B & A & HM \\
		\midrule[0.5pt]
		1 & \underline{30.0} & \textcolor{gray}{64.4} & \textcolor{gray}{57.5} & \underline{40.9} \\
		2 & \textbf{30.1} & \textcolor{gray}{64.4} & \textcolor{gray}{57.5} & \textbf{41.0} \\
		3 & 29.4 & \textcolor{gray}{64.3} & \textcolor{gray}{57.3} & 40.4 \\
		\bottomrule[0.8pt]
	\end{tabular}
	\caption{Study on the number of class prototypes during inference.}
	\label{table4}
\end{table}
Table 3 demonstrates the effectiveness of the dynamic score aggregation strategy in addressing the issue where multiple prototypes represent the same category during the inference phase. The experimental results show that selecting two class prototypes for dynamic aggregation of discriminant scores achieves the best detection performance for unknown categories, compared to using a single prototype or three prototypes. This mechanism improves discriminative robustness in open-vocabulary scenarios through collaborative decision-making among prototypes \cite{2}.

\section{D Open-vocabulary Aerial Object Detection Benchmark Details}
\begin{table*}[ht]
	\centering
	\renewcommand{\arraystretch}{0.8}
	\small
	\begin{tabular}{c|c|ccc|ccc}
		\hline
		\multicolumn{1}{c|}{}& \multicolumn{1}{c|}{} & \multicolumn{3}{c|}{\textbf{\# Images}} & \multicolumn{3}{c}{\textbf{\# Categories}} \\
		\multirow{-2}{*}{\textbf{Dataset}} & \multirow{-2}{*}{\textbf{Images resolutions}} & train & val & total & novel & base & all \\
		\hline
		\multicolumn{8}{c}{\textbf{Original Dataset}} \\
		\hline
		DIOR & $800 \times 800$ & $5862 (all)$ & $5863$ & $11725$ & $0$ & $0$ & $20$ \\
		DOTAv1.0 & $800 \times 800 \sim 4000 \times 4000$ & $1141 \rightarrow 10360 (all)$ & $458 \rightarrow 3237$ & $1599 \rightarrow 13597$ & $0$ & $0$ & $15$ \\
		\hline
		\multicolumn{8}{c}{\textbf{Preprocessed Dataset}} \\
		\hline
		DIOR & $800 \times 800$ & $4933 (base)$ & $5863$ & $10796$ & $4$ & $16$ & $20$ \\
		DOTAv1.0 & $800 \times 800$ & $ 9692 (base) $ & $3237$ & $12929$ & $4$ & $11$ & $15$ \\
		\hline
	\end{tabular}
	\caption{A summary of the datasets employed in our experiments. $\rightarrow$ indicates data preprocessing steps, including resolution-normalized cropping and quality filtering of unlabeled images. $(all)$ and $(base)$ denote subsets of the data: $(all)$ includes all categories (both base and novel), while $(base)$ includes only the base categories.}
	\label{tab:dataset_specs}
\end{table*}

To ensure experimental fairness, we conduct our experiments under OVAD setting established by DescReg \cite{3}. We assess VK-Det's capability in detecting novel objects on two benchmark datasets, with detailed dataset information provided in Table 4.
\begin{itemize}
	\item DIOR is a large benchmark dataset in the field of AOD, with uniformly sized images of 800×800 pixels. It contains 5,862 training images, 5,863 validation images, and 11,725 test images, covering 20 labeled object categories. 
	
	During VK-Det training, we applied a filtering strategy to remove all annotated samples from novel categories in the training set, resulting in 4,933 valid training images. This step ensures that the model does not encounter semantic information related to novel categories during training. Notably, since the pseudo-labeling mechanism does not depend on category labels, we further fine-tuned the model using the original set of 5,862 training images as unlabeled data, while keeping the validation set unchanged to evaluate the model's generalization ability.
	\item DOTA consists of satellite images with resolutions ranging from 800×800 to 4000×4000 pixels. In this study, we use the DOTAv1.0 dataset, which contains 1,411 images in the original training set and 456 images in the validation set (all with variable resolutions). 
	
	To unify the input scale, we follow the standard preprocessing procedure \cite{3} to crop the images: first, we expand the training set to 18,430 images and the validation set to 6,259 images (all with 800×800 pixels); and then, based on the principle of quality filtering, we further filter out 8,070 training images and 3,022 validation images. To strictly avoid exposure to novel categories during model training, two key measures are adopted: (i) only 9,692 labeled images containing the base categories are retained in the training set; and (ii) an additional 10,360 unlabeled images in the training set are used for pseudo-labeled data generation. The validation set maintains the original 3,237 images to ensure the generalization of the evaluation results.
\end{itemize}

We followed the category division proposed in previous work \cite{3}, categorizing the DIOR dataset into 16 base and 4 novel categories, and the DOTA dataset into 11 base and 4 novel categories.
\begin{itemize}  
	\item DIOR:  
	\begin{itemize}  
		\item Base categories: 'airplane', 'baseball field', 'bridge', 'chimney', 'dam', 'Expressway Service Area', 'Expressway Toll Station', 'golf field', 'harbor', 'overpass', 'ship', 'stadium', 'storage tank', 'tennis court', 'train station'.  
		\item Novel categories: 'vehicle', 'airport', 'basketball court', 'ground track field', 'windmill'.  
	\end{itemize}  
	\item DOTA:  
	\begin{itemize}  
		\item Base categories: 'plane', 'ship', 'storage tank', 'baseball diamond', 'basketball court', 'ground track field', 'harbor', 'bridge', 'large vehicle', 'small vehicle', 'roundabout'.  
		\item Novel categories: 'tennis court', 'helicopter', 'soccer ball field', 'swimming pool'.  
	\end{itemize}  
\end{itemize}  

\section{E Training Details of the Relevant Models}
\subsection{E.1 Details of the VK-Det Framework}

\textbf{OLN training details.} 

We use the OLN model \cite{4} based on the Resnet50-FPN architecture to pre-train the class-agnostic region proposal generator. The model is pre-trained on a filtered training set (without novel category labels) using a single A800 GPU, executing 8 epochs (1x schedule) with a batch size of 8. After training, this OLN model provides three types of key data support for VK-Det: 1) proposal data for the filtered training set; 2) proposal data for the unlabeled training set; and 3) proposal data for the validation set. All of these proposal data have precise localization attributes and remain class-agnostic, laying the foundation for subsequent detection tasks.

\textbf{VK-Det training details.}

The VK-Det framework is built upon the LP-OVOD \cite{8} configuration, with its ResNet50 backbone initialized using a self-supervised pre-trained SoCo model \cite{5}. During the AFKD training phase, we employ the SGD optimizer with a base learning rate of 0.01 on a single A800 GPU, and the learning rate is reduced to one-tenth of its initial value at the 16th and 19th epochs. In the subsequent PAPL fine-tuning phase, a fine-tuning dataset is constructed by selecting 500 high-quality pseudo-labels per unknown category. The prototype classifier is then fine-tuned for 12 epochs, with the learning rate of 0.001 reduced to one-tenth of its initial value at the 8th and 11th epochs. During the testing phase, the temperature coefficients of both the distillation head and classification head are uniformly set to 0.01 to enhance prediction confidence of novel categories.

\subsection{E.2 Details of the Compared Methods}

\textbf{ViLD training details.} 

The original ViLD model is trained end-to-end on multiple TPU devices with a batch size of 256, for a total of 180,000 iterations \cite{6}. Input images are uniformly scaled to a resolution of 1024×1024 and undergo large-scale jitter augmentation. To ensure a fair comparison with ViLD, we strictly follow its paper and publicly available code to reproduce the training procedure. Concretely, class-agnostic proposals are generated using the OLN model, and the top 300 proposals with the highest confidence scores from each image are selected for training. The model is trained for 20 epochs using the same optimizer configuration as the AFKD module. All other hyper-parameters are kept consistent with our ASKD experimental setup.

\textbf{CastDet training details.} 

Strictly following the CastDet paper and its open-source code specifications \cite{1}, we first pre-trained Soft Teacher \cite{7} on the filtered dataset (masked with novel category labels). The training was conducted for 8,000 iterations with a batch size of 12 using a single A800 GPU. The resulting Soft Teacher weights were then fused with the pre-trained RemoteCLIP-R50 model \cite{8}. Subsequently, the unlabeled training set was utilized as unlabeled data for fine-tuning, which was performed for 10,000 iterations with a batch size of 24 using the SGD optimizer (learning rate: 0.01, momentum: 0.9, weight decay: 0.0001). All hyper-parameters remained consistent with those specified in the original literature and code implementation.

\section{F Results of the Qualitative Analysis}

\subsection{F.1 Visualization of Informative Region Perception}
\begin{figure*}[t]
	\centering
	\includegraphics[width=1\textwidth]{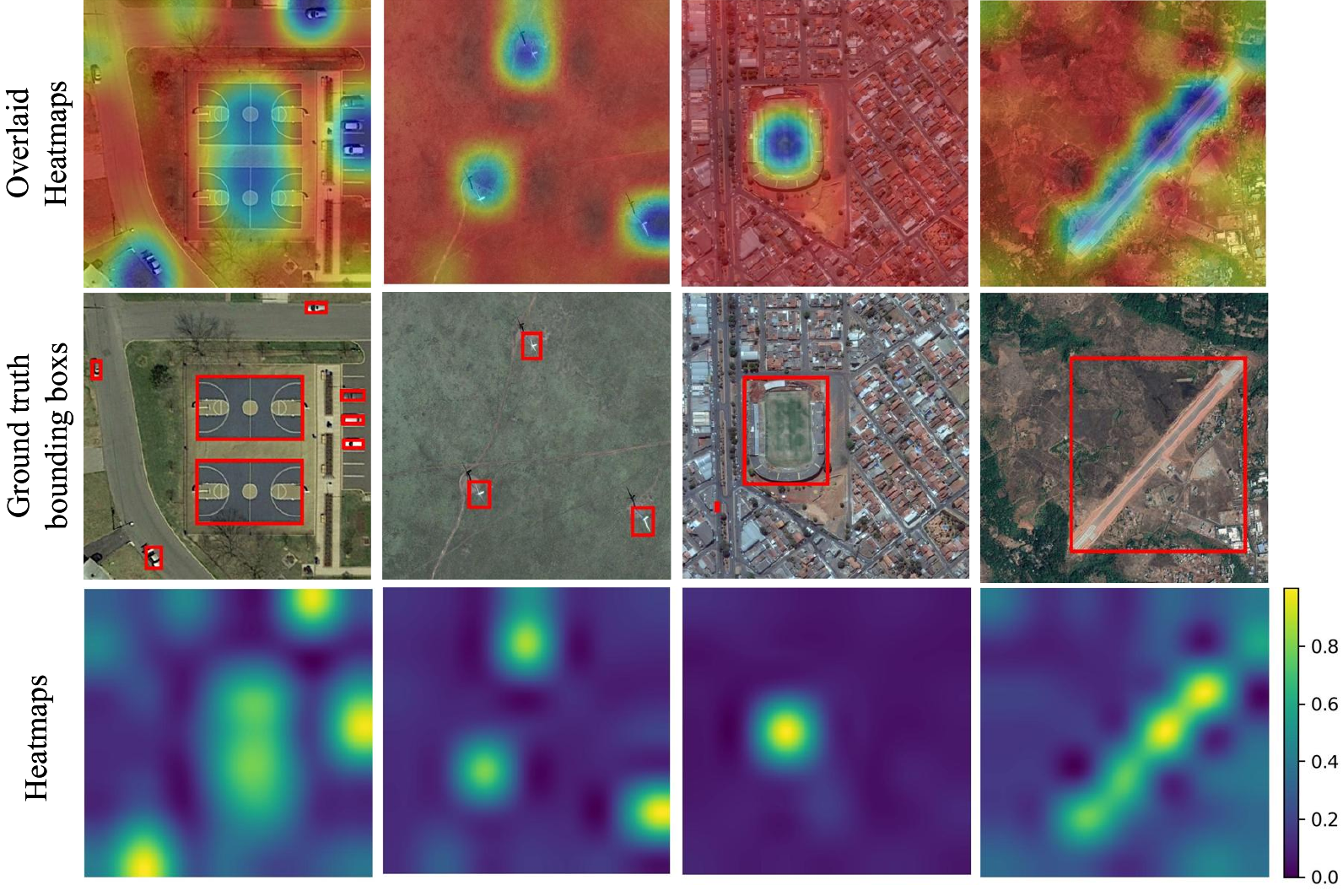} 
	\caption{Visualization of informative region perception.}
	\label{fig6}
\end{figure*}

Fig.~\ref{fig6} illustrates the high degree of spatial consistency between the heatmap of informative region perception and the ground truth bounding boxes. This is demonstrated through three visual comparisons of the same image: a multi-layer average attention heatmap based on the visual encoder of VLMs, the ground truth bounding boxes, and the heatmap overlaid on the original image. This visualization effectively supports the feasibility of the VK-Det framework.

\subsection{F.2 Visualization of Different Pseudo-labeling Methods}

\begin{figure*}[t]
	\centering
	\includegraphics[width=1\textwidth]{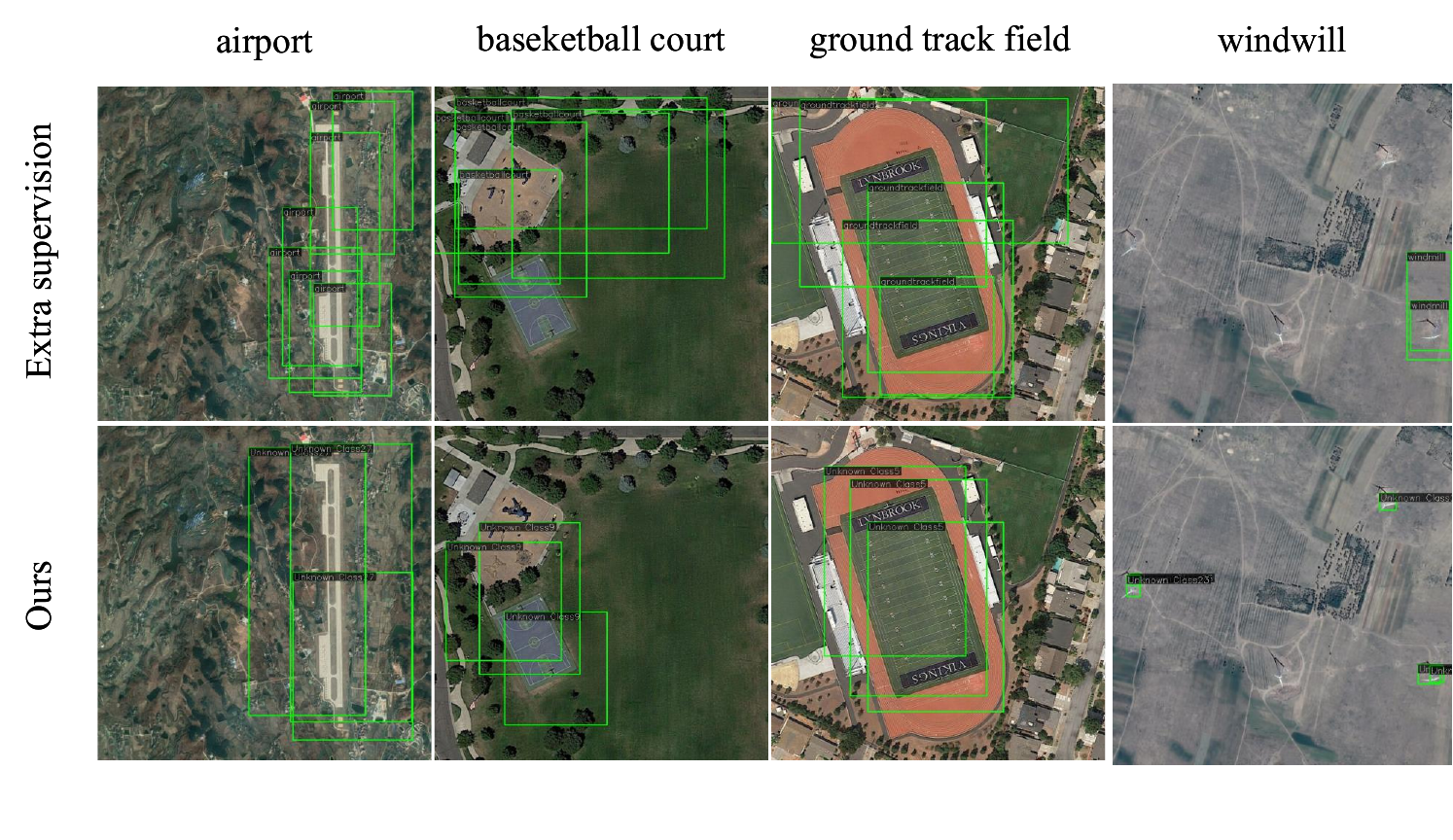} 
	\caption{Visualization of different pseudo-labeling methods.}
	\label{fig7}
\end{figure*}
Fig.~\ref{fig7} illustrates the qualitative analysis comparing the two pseudo-label generation methods: clustering-based pseudo-label generation extracts image embeddings of proposals through the visual encoder of VLMs, and the 100 image embeddings with the closest distance to the cluster centers are selected to construct the pseudo-labels; whereas text-supervised pseudo-label generation generates text embeddings using a fixed-category descriptive template, and then selects the image embeddings with the highest similarity among the 100 proposals. The visualization results show that the proposed method can effectively suppress text noise interference and generate higher-quality pseudo-labeled data while reducing background interference.

\subsection{F.3 Visualization of Inference}
\begin{figure*}[t]
	\centering
	\includegraphics[width=1\textwidth]{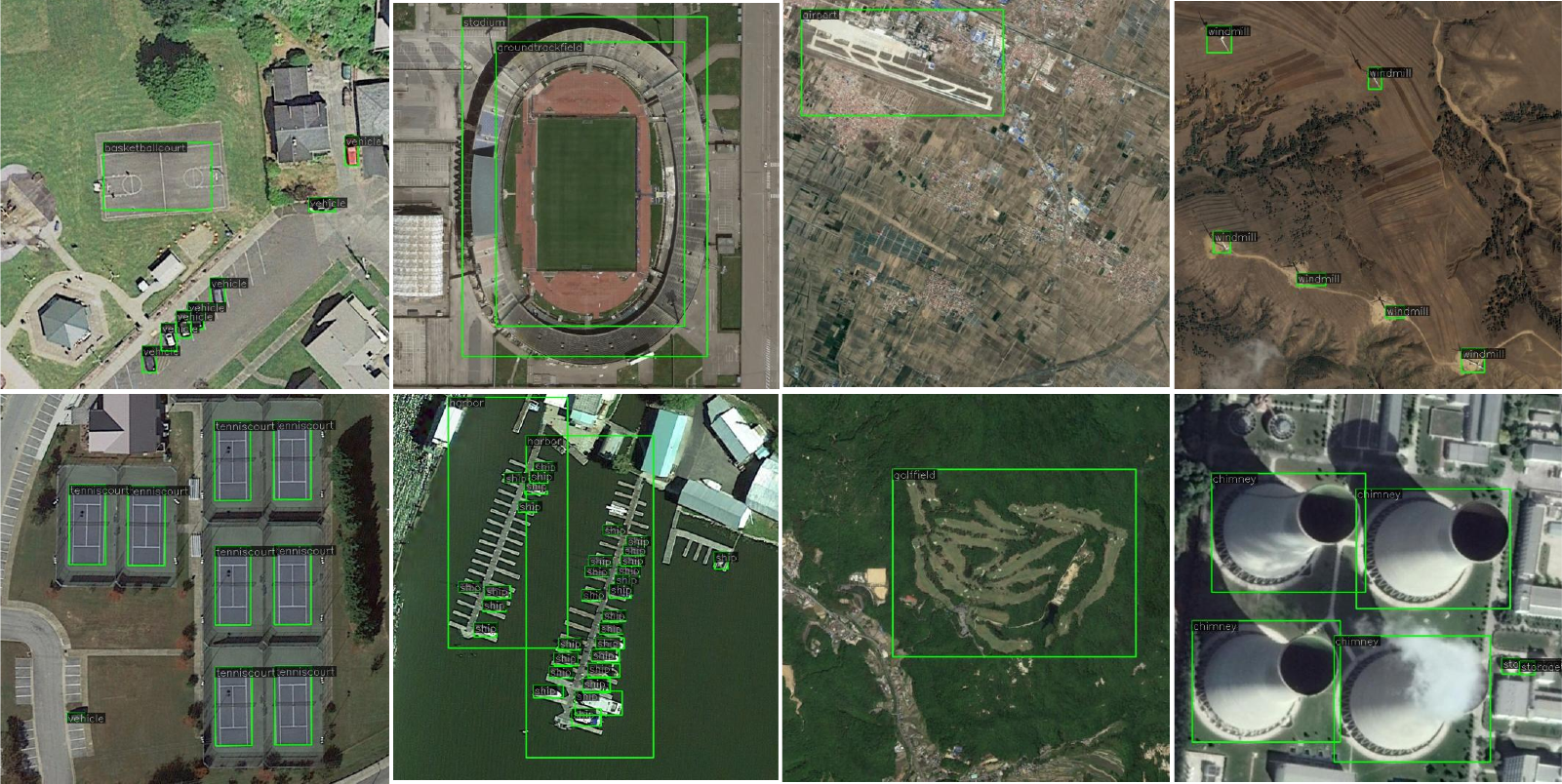} 
	\caption{Visualization of inference.}
	\label{fig8}
\end{figure*}
Fig.~\ref{fig8} shows the inference results of VK-Det on the DIOR dataset, indicating the effectiveness of our method in open-vocabulary aerial object detection.

\end{document}